\documentclass[times, review, 10pt]{elsarticle}
 
\usepackage{amssymb}
\usepackage{array}
\usepackage{graphicx}
\usepackage{picinpar}
\usepackage{amsmath}
\usepackage{url}
\usepackage{flushend}
\usepackage{colortbl}
\usepackage{soul}
\usepackage{multirow}
\usepackage{amsmath}
\usepackage{amssymb}
\usepackage{pifont}
\usepackage{graphicx}
\usepackage{color}
\usepackage{adjustbox}
\usepackage[hidelinks]{hyperref}
\usepackage{enumerate}
\usepackage{siunitx}
\usepackage{breakurl}
\usepackage{epstopdf}
\usepackage{pbox}
\usepackage{enumitem}
\usepackage{subcaption}  
\usepackage{multirow}
\usepackage{booktabs}
\usepackage{multirow}
\usepackage{makecell}
\usepackage{array}
\usepackage{varioref}
\usepackage{caption}
\usepackage{bbold}
\usepackage{amsmath,amssymb}
\usepackage{orcidlink}
\usepackage{scalerel}
\usepackage{tikz}
\usepackage{comment}
\usepackage{ifsym}

\journal{Pattern Recognition}

\begin{document}

\begin{frontmatter}



\title{DyConfidMatch: Dynamic Thresholding and Re-sampling for 3D Semi-supervised Learning}

\author[]{Zhimin Chen}
\author[]{Bing Li*}
\fntext[]{DISTRIBUTION STATEMENT A. Approved for public release; distribution is unlimited. OPSEC 8735}
\ead{bli4@clemson.edu}
\affiliation[]{organization={Clemson University},
            addressline={4 Research Drive},
            city={Greenville},
            postcode={29607},
            state={SC},
            country={USA}}

\begin{abstract}

Semi-supervised learning (SSL) leverages limited labeled and abundant unlabeled data but often faces challenges with data imbalance, especially in 3D contexts. This study investigates class-level confidence as an indicator of learning status in 3D SSL, proposing a novel method that utilizes dynamic thresholding to better use unlabeled data, particularly from underrepresented classes. A re-sampling strategy is also introduced to mitigate bias towards well-represented classes, ensuring equitable class representation. Through extensive experiments in 3D SSL, our method surpasses state-of-the-art counterparts in classification and detection tasks, highlighting its effectiveness in tackling data imbalance. This approach presents a significant advancement in SSL for 3D datasets, providing a robust solution for data imbalance issues.

\end{abstract}




\end{frontmatter}


\section{Introduction}
The collection and annotation of 3D data is a costly and labor-intensive process. As a result, 3D semi-supervised learning (SSL) has gained significant interest recently, given its efficacy in utilizing unlabeled data \cite{wang20213dioumatch,chen2021multimodal,sohn2020fixmatch,zhang2021flexmatch}. Many SSL methods, including Pseudo-Labeling\cite{lee2013pseudo} and FixMatch~\cite{sohn2020fixmatch}, employ the pseudo-labeling strategy, which leverages high-confidence predictions of unlabeled data as labels to further improve network performance.

\begin{figure}[t!]
    \centering
    \begin{subfigure}[b]{0.45\textwidth}   
        \centering 
\includegraphics[width=6cm,height=5.5cm]{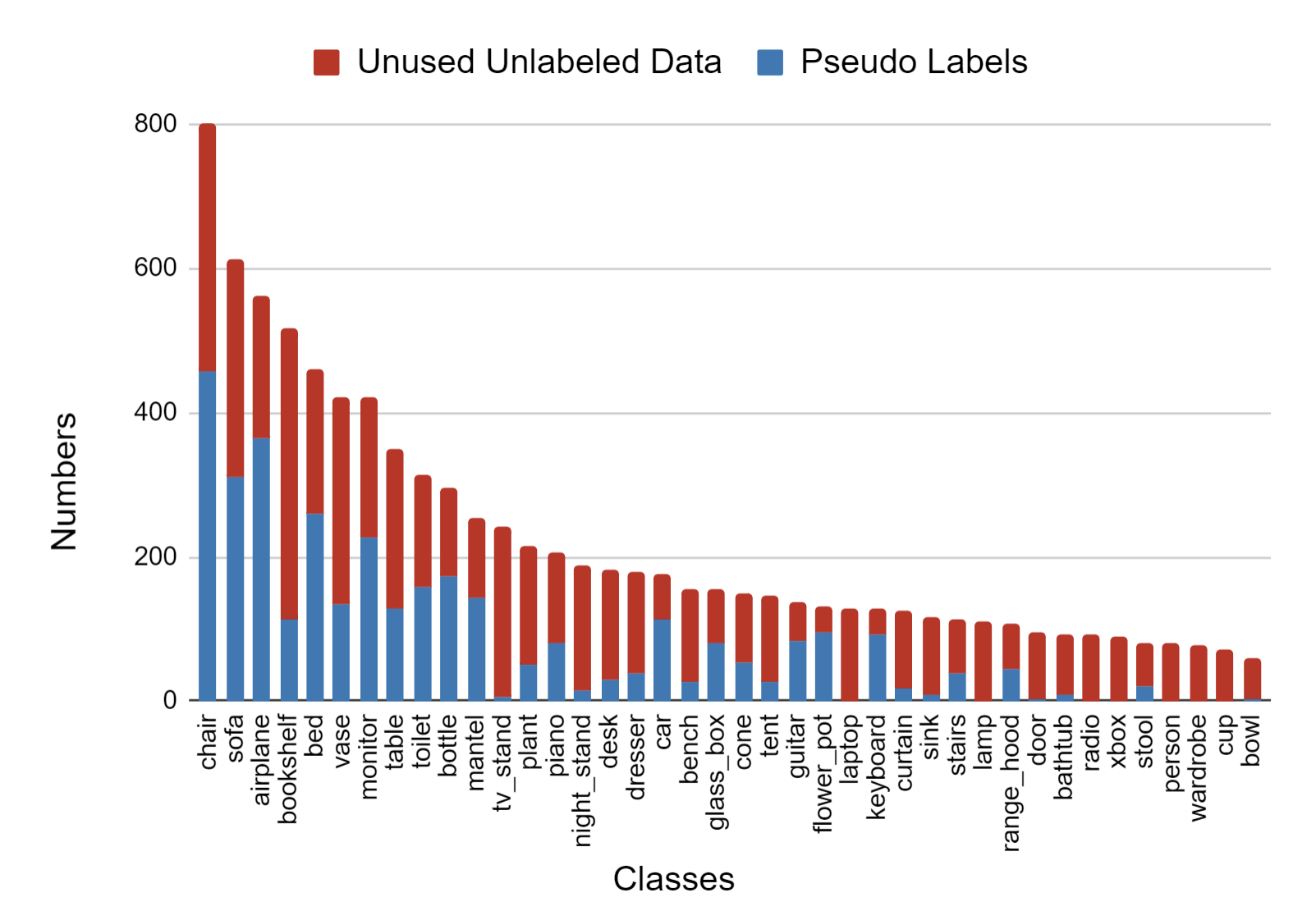}
        \caption[]%
        {{\small }}    
        \label{fig:th07}
    \end{subfigure}
    \begin{subfigure}[b]{0.45\textwidth}   
        \centering 
        \includegraphics[width=6cm,height=5.5cm]{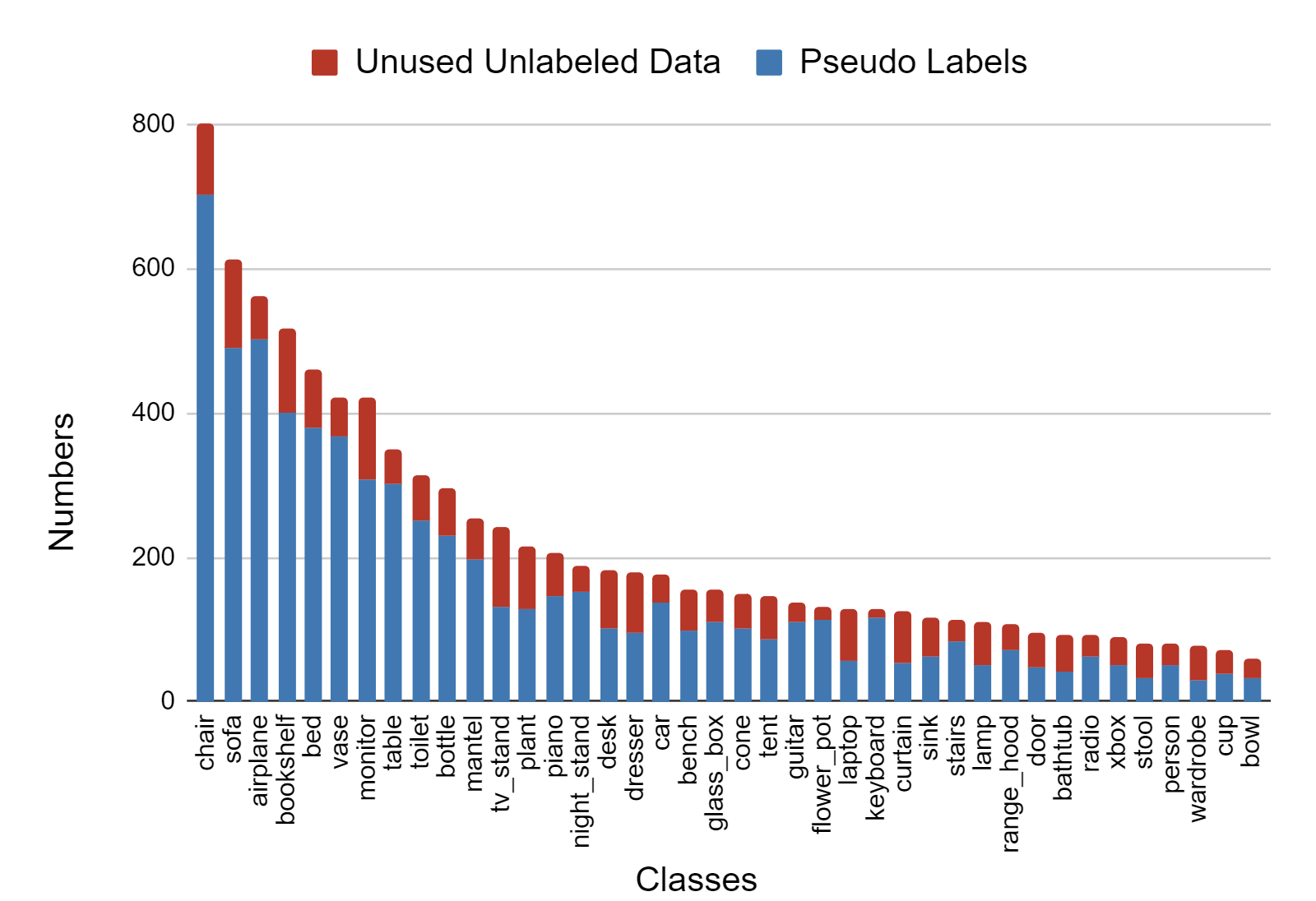}
        \caption[]%
        {{\small }}    
        \label{fig:th03}
    \end{subfigure}

    \caption[ ]
    {\small {The analysis of FixMatch, when trained on the ModelNet40 dataset using only 10\% labeled data, reveals the following insights based on fixed threshold values of 0.9 and 0.3: (a) The number of pseudo-labels selected from the unlabeled data varies significantly. With a higher threshold of 0.9, only a small portion of the unlabeled data is incorporated, as the stricter criterion filters out many potential pseudo-labels. (b) Conversely, with a lower threshold of 0.3, a larger portion of the unlabeled data is utilized, as more pseudo-labels are accepted. Therefore, to fully leverage the unlabeled data, a relatively lower threshold should be considered.} }
    \label{fig:acc-confid-relationship}
\end{figure}

Despite their widespread use and demonstrated performance improvements across various tasks~\cite{zhao2020sess,wang20213dioumatch,chen2021multimodal,sohn2020fixmatch, yang2024information, huang2022assister}, fixed pseudo-labeling methods have an inherent limitation: they depend heavily on a predetermined threshold to filter low-quality pseudo-labels. In this approach, any unlabeled data point with confidence higher than the threshold is utilized for training, while the rest is discarded, regardless of the category. To ensure high-quality pseudo-labels, most implementations adopt a high threshold (e.g., 0.9)\cite{wang20213dioumatch,chen2021multimodal} in 3D tasks.

\begin{figure}[t!]
    \centering
    \begin{subfigure}[b]{0.45\textwidth}   
        \centering 
\includegraphics[width=5.5cm,height=5cm]{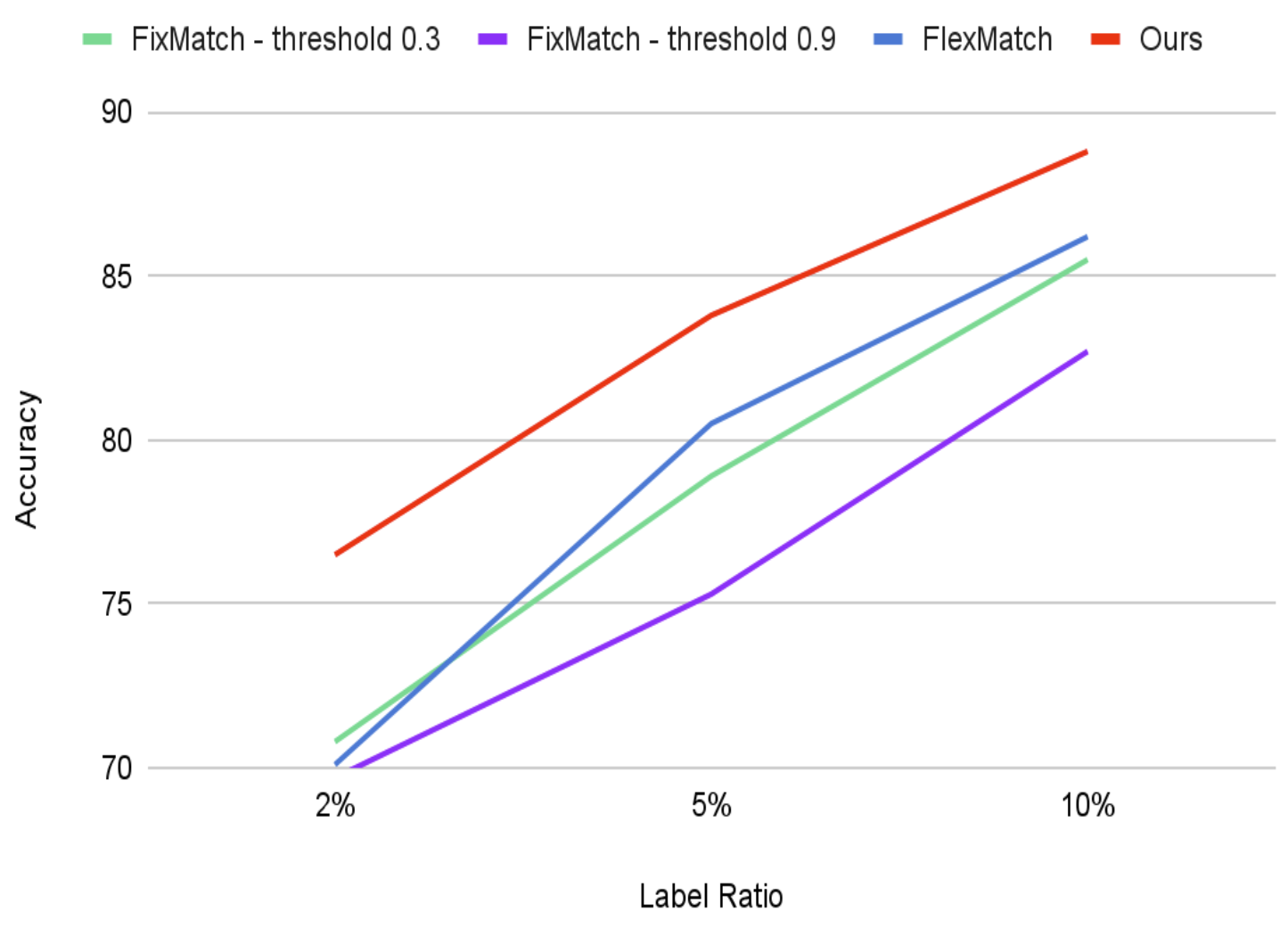}
        \caption[]%
        {{\small }}    
        \label{fig:compare_acc}
    \end{subfigure}
    \begin{subfigure}[b]{0.45\textwidth}   
        \centering 
        \includegraphics[width=5.5cm,height=5cm]{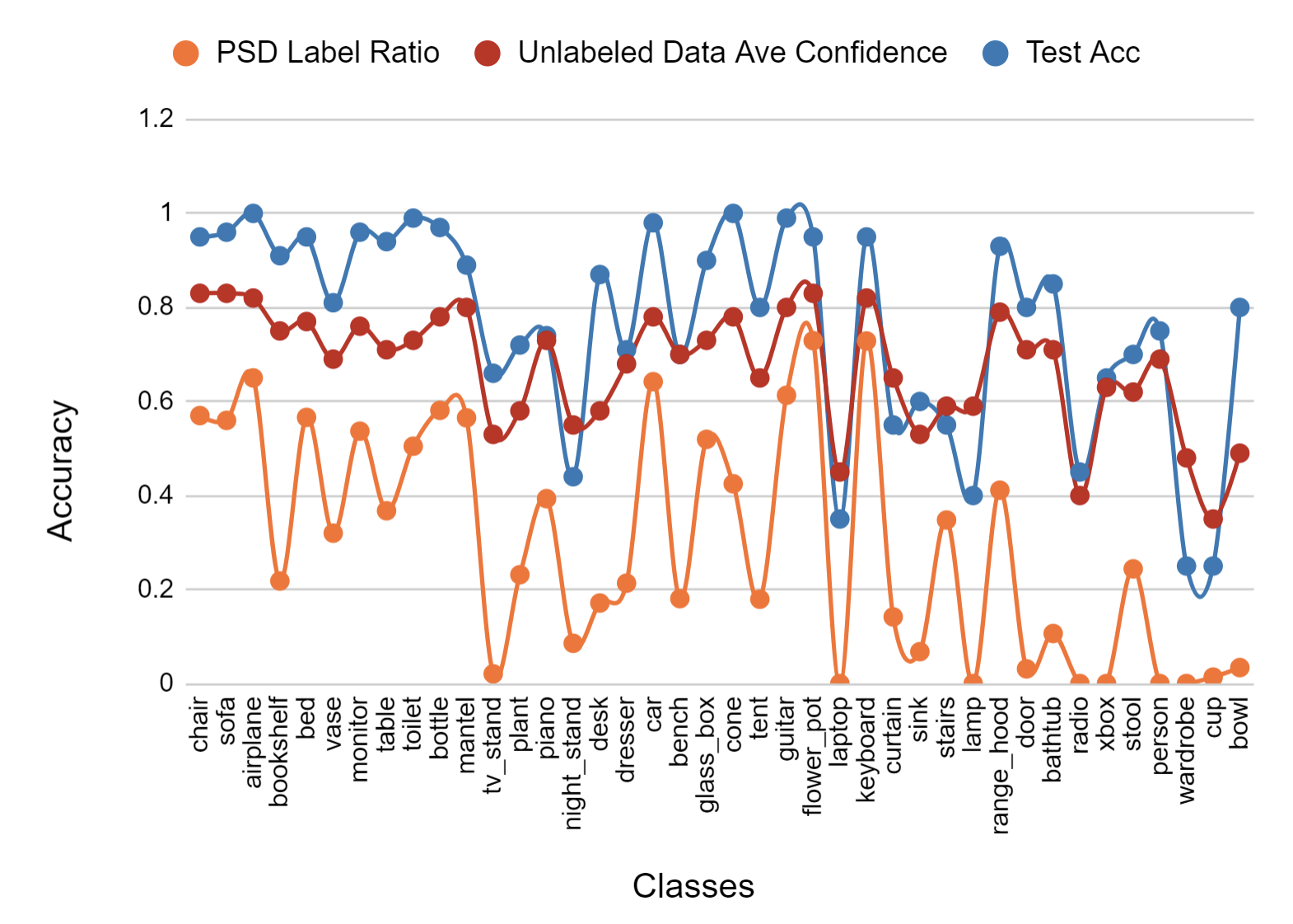}
        \caption[]%
        {{\small }}    
        \label{fig:modelnet40-confid}
    \end{subfigure}

    \caption[ ]
    {\small {(a)The results comparing FixMatch with 0.3 and 0.9 thresholds, FlexMatch, and our method show that while a lower threshold (0.3) allows for greater utilization of unlabeled data, it also introduces more noise, which negatively impacts performance. In contrast, our method strikes a balance between pseudo-label quality and the effective use of unlabeled data, leading to superior overall performance. (b) There are notable differences in the test accuracy and the confidence level for each class in the unlabeled dataset. The learning challenge varies widely across different classes, and intriguingly, some less common classes register a better accuracy than the more prevalent ones. Furthermore, (1) The limited utilization of unlabeled data with a high threshold suggests a cautious approach to incorporating such data during training. (2) The observed differences in learning difficulty among classes highlight the importance of addressing class imbalance during training. (3) The strong correlation between class-level confidence and test accuracy underscores the reliability of the confidence scores in predicting class performance.} }
    \label{fig:acc-confid-relationship}
\end{figure}

While these methods have demonstrated strong performance across various tasks, they have an inherent limitation: reliance on a fixed threshold to filter pseudo-labels~\cite{zhao2020sess, wang20213dioumatch}. Specifically, only unlabeled data points with confidence higher than the threshold are used for training, while the rest are discarded. This approach, often adopting a high threshold (e.g., 0.9)~\cite{wang20213dioumatch}, works well in 2D tasks where predictions tend to be more confident and stable. However, in 3D tasks, where confidence levels are generally lower, such high-threshold strategies lead to the underutilization of unlabeled data, particularly from underrepresented classes (Fig.~\ref{fig:th07}). Conversely, As shown in Fig.~\ref{fig:th03}, lowering the threshold (e.g., to 0.3) allows more unlabeled data to be utilized but introduces noise in the form of low-quality pseudo-labels, which ultimately degrades model performance (Fig.~\ref{fig:compare_acc}). This problem is further worsened by the class imbalance inherent in many 3D datasets, where certain classes are overrepresented while others remain scarce (Fig.~\ref{fig:th07}). 

To address these challenges, a specialized approach is necessary—one that is explicitly designed for the 3D modality and takes into account the unique characteristics of 3D data, such as lower confidence levels and class imbalances, which are often overlooked by 2D counterpart methods \cite{xiang2023hybrid, xiang2024imitation, zhang2021flexmatch, sohn2020fixmatch, lee2013pseudo, yao2023goal, deng2023long, deng2023plgslam, deng2024compact, zhao2020u}. In this work, we aim to develop a novel method for assessing the learning status of each class and propose a class-level pseudo-label selection mechanism that balances both the quality and quantity of pseudo-labels. Previous studies \cite{lee2013pseudo, sohn2020fixmatch} have utilized instance-level confidence to gauge learning status in imbalanced datasets, proving effective for pseudo-label selection. Building on this concept, we introduce class-level confidence to represent learning status at the class level. Estimating each class's learning status through class-level confidence provides a more accurate reflection of the model's proficiency with each class, as evidenced by our analysis (Fig.~\ref{fig:modelnet40-confid}).

Building on our intuitive insights and analysis, a semi-supervised learning method is proposed, which is capable of dynamically adjusting the threshold based on the learning status. Our method dynamically adjusts the threshold based on each class's learning status, thereby allowing greater utilization of unlabeled data from those classes with a lower learning status. Furthermore, more unlabeled data will be used in our method during the initial training stage, particularly when limited data have predicted confidence exceeding the static high thresholds. However, it is pertinent to mention that while increasing the count of selected pseudo-labels through dynamic thresholding does improve matters, it does not completely eradicate the learning status variance resulting from data imbalance and learning difficulty variance. Such a situation can lead the network to exhibit a bias towards classes with a higher learning status, consequently leading to overfitting.

To circumvent this issue, we leverage the class-level learning status to dynamically sample the dataset. Specifically, this strategy enhances the sampling probability of instances from classes with lower learning status, while reducing it for those classes with higher learning status. By combining dynamic thresholding and re-sampling, our method balances the learning status of all classes and effectively leverages the unlabeled data. The aim of our approach is to assess each class's learning status and subsequently refine and balance it. When compared with other 3D SSL methodologies, our technique stands out due to its ability to estimate each class's learning status, dynamically adjust the threshold, and re-sample data. Unlike FlexMatch~\cite{zhang2021flexmatch}, which primarily focuses on learning difficulty, our approach also incorporates data imbalance, ensuring a more balanced and effective learning process. In this work, we define learning status as the degree of proficiency with which the model learns a class, as indicated by test accuracy. We argue that both learning complexity and data imbalance significantly influence a network's learning status, and our method accounts for both factors. Our approach can be applied to a variety of semi-supervised tasks to enhance performance, even in the context of imbalanced datasets.

To illustrate the broad applicability of this work, we have conducted evaluations on both SSL 3D object recognition and SSL 3D object detection tasks. The results clearly demonstrate our method's superior performance over existing state-of-the-art techniques.

Key contributions of our method are as follows: 
\begin{enumerate}[itemindent=1em]
    \item 
    This work provides a holistic definition of learning status, which not only considers learning difficulty but also considers the effects of imbalanced data. Furthermore, we empirically indicate that class-level confidence reflects the learning status of individual classes.

    \item 
    We present an innovative 3D SSL method that simultaneously tackles imbalanced data and learning difficulty issues according to learning status. Our method dynamically adjusts thresholds and re-samples data depending on each class's learning status, balancing learning status, and effectively addressing both challenges.

    \item 
    This research notably surpasses current SOTA 3D SSL object detection and classification techniques, demonstrating its efficiency.
\end{enumerate}

\textbf{Key Distinctions from Our Conference Paper:} This article has substantial improvements over the conference version~\cite{chen2023class}: 
(i) A new dynamic max thresholding method is proposed based on class-level confidence to reduce the dependency on the manner of setting threshold value and further improve the results. (ii) We explore the correlation between learning status and class-level confidence in 3D SSL detection tasks, which substantiates the universality of our hypothesis. (iii) To demonstrate the generality of this work, we implement the proposed method on the outdoor object detection dataset (KITTI~\cite{geiger2013vision}). (iv) This work significantly outperforms the previous conference version in both 3D object classification and detection tasks.

\section{Related work}

\subsection{Semi-Supervised Learning} Over recent years, SSL methodologies have demonstrated remarkable progress ~\cite{xu2021dash, chen2023bridging, feng2022advancing, feng2022dmt, yao2024event, xin2024let, Fu2024DetectingMI}. Pseudo-labeling ~\cite{lee2013pseudo} method is leveraged by many contemporary SSL methods to minimize the predictions' uncertainty on unlabeled data. Therefore, the effectiveness SSL methods is largely dependent on the quality of the pseudo-labels. Fang et al. ~\cite{fang2020collaborative} proposes to cluster deep features to generate pseudo labels for abundant unlabeled samples. To enhance pseudo-labels' quality, the leading-edge SSL method FixMatch~\cite{sohn2020fixmatch} and various similar approaches typically implement a high, fixed threshold to eliminate predictions that have low confidence scores emerging from strongly augmented data. While this high-threshold tactic enhances pseudo-label quality, it overlooks the learning status of individual classes. This results in a bias towards classes with higher learning status and makes a significant amount of unlabeled data unused. In response to this problem, Dash~\cite{xu2021dash} proposes a dynamic thresholding method for all classes based on cross-entropy loss. Meanwhile, Based on curriculum learning, the FlexMatch~\cite{zhang2021flexmatch} replaces the rigid, predefined threshold with a flexible threshold, thereby taking into account the class-level learning status. However, our empirical analyses suggest that FlexMatch does not adapt well to 3D datasets. This is because the FlexMatch is designed for class-balanced datasets, whereas labels in each class in the 3D datasets are unbalanced. Moreover, the prediction confidence for certain classes from the 3D networks fails to reach the predefined high threshold, resulting in significantly reduced class thresholds. Thus, the strategy to estimate learning status in FlexMatch proves unsuitable for 3D SSL tasks.

 \subsection{Class-Imbalanced Semi-Supervised Learning}
Class-imbalanced SSL gained traction recently, as it provides a more accurate representation of the data distribution~\cite{yang2020rethinking,hyun2020class, Fu2024DetectingMI, lai2023xvo}. Wei et al.~\cite{wei2021crest} observed that conventional SSL methods typically exhibit high recall but low precision for head classes. To address this, they introduced CReST, which uses the quantities of labeled data to re-samples unlabeled data. The recent cutting-edge method BiS~\cite{he2021rethinking} employs two distinct re-sampling strategies concurrently to train the model in a decoupled manner. All these class-imbalanced SSL approaches re-sample based on data quantities. Nevertheless, based on our observations, it has become evident that specific minority classes possess the potential to surpass the performance of majority classes, primarily because they are inherently easier to learn. If re-sampling is solely based on the volume of data, it might inadvertently bias the model in favor of these classes that present a lower learning challenge. Thus, it's crucial to consider not just the quantity but also the intrinsic difficulty of each class when re-sampling, ensuring that the model is well-balanced and robust.


\begin{figure*}[t!]
\centering
\setlength{\belowcaptionskip}{-10pt}
\includegraphics[width=12cm,height=5.3cm]{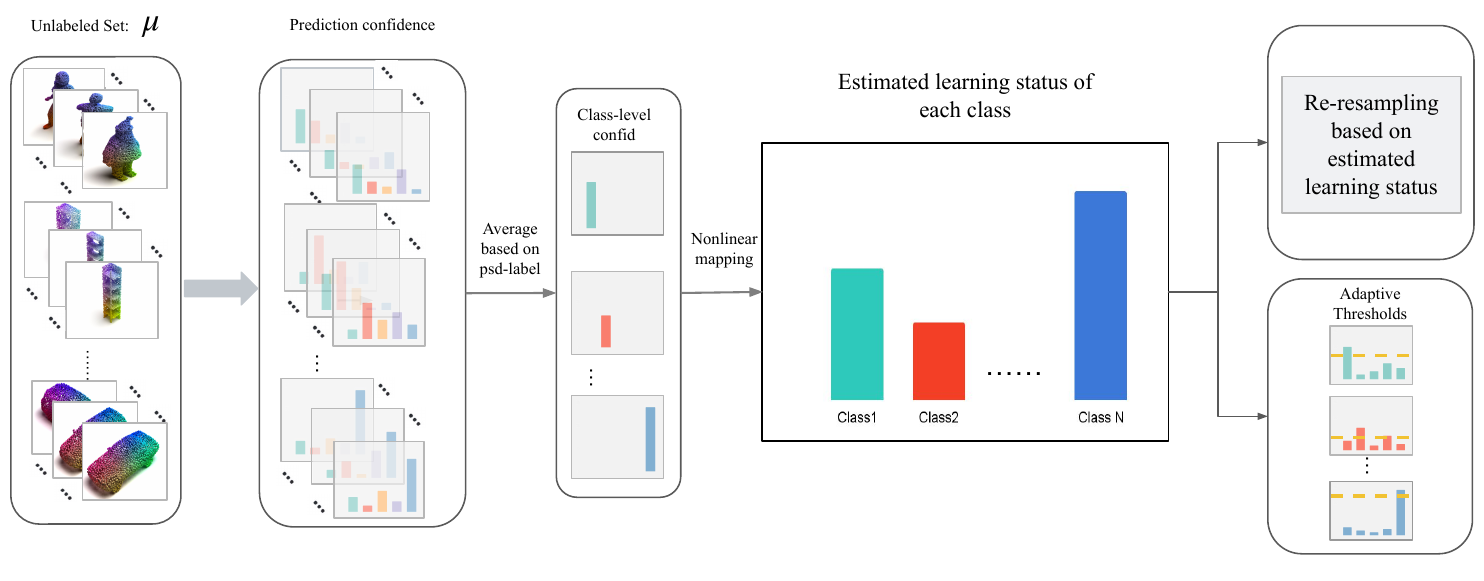}
\caption{The proposed method consists of three key components: (1) Leveraging class-level confidence derived from unlabeled data to determine the learning status for each class, (2) Utilizing the learning status to dynamically adjust the threshold for each class, and (3) Adjusting the re-sampling strategy based on the identified learning status.}
\label{fig:framework}
\end{figure*}

\subsection{3D Semi-Supervised Learning} Despite 3D understanding tasks have achieved noteworthy results, these methods often require high-quality 3D labels, which are costly to collect. Given the ability of SSL to reduce reliance on labeled data, SSL applied to 3D object detection has attracted considerable research interest ~\cite{wang20213dioumatch, chen2021multimodal, li2023learning, lin2023self}. Recently, Yin et al. ~\cite{yin2022semi} introduced a novel teacher-student pseudo-labeling framework for semi-supervised 3D object detection. This framework leverages the strengths of teacher-student models to enhance the accuracy and robustness of pseudo-labels in 3D object detection tasks. Ssda3d ~\cite{wang2023ssda3d} explored intra-domain point-MixUp techniques in semi-supervised learning, effectively regularizing the pseudo-label distribution and improving model generalization. ProposalContrast ~\cite{yin2022proposalcontrast} proposed a method to learn robust 3D representations by contrasting region proposals. This approach emphasizes the importance of differentiating between various 3D regions, thereby enhancing the discriminative power of the learned representations. Similarly, Dqs3d ~\cite{gao2023dqs3d} introduced a framework for densely-matched quantization-aware semi-supervised 3D object detection, addressing the challenges of quantization in semi-supervised settings and improving detection accuracy. Moreover, Diffusion-SS3D ~\cite{ho2024diffusion} enhanced the quality of pseudo-labels through the diffusion model for semi-supervised 3D object detection. This method utilizes diffusion processes to refine pseudo-labels, leading to more accurate and reliable detection outcomes. Together, these advancements represent significant strides in the field of semi-supervised 3D object detection, each contributing unique techniques and insights to improve the efficacy and robustness of 3D object detection models. However, most existing 3D SSL detection methodologies select pseudo-labels with a fixed threshold like FixMatch. Many existing strategies overlook the variance of learning difficulty and the complexities arising from class imbalances, often leading to sub-optimal results. In contrast, our approach introduces an innovative 3D SSL technique rooted in class-level confidence assessments. This advanced method facilitates adaptive threshold adjustments and data re-sampling, tailored to the current learning state of the dataset. By doing so, it offers a promising solution to the challenges previously highlighted, ensuring a more comprehensive and effective learning process.

\section{Method}

An outline of our proposed method is depicted in Figure~\ref{fig:framework} and it has three key parts: (1) Learning status estimation, which leverages prediction confidence from unlabeled data to generate class-level confidence and thus obtain the learning status. (2) Dynamic thresholding, implemented based on each class's learning status. (3) Dynamic re-sampling, executed individually for each class, taking into account their learning status. Detailed explanations and formulations for each of these components are discussed in the following.

\subsection{Problem Formulation}
The primary objective of 3D SSL is to train models by utilizing limited labeled samples in conjunction with a substantial quantity of unlabeled samples. Let's denote $X_L = {(x_i,y_i)}{i=1}^{N_L}$ as a labeled dataset with a relatively small sample size of $N_L$. Here, $x_i\in \mathbb{R}^{N\times3}$ represents a 3D point cloud depiction of an object or scene, with $y_i$ as the associated label. We also define $X_U = {(x_i)}{i=1}^{N_U}$ as an unlabeled dataset comprising $N_U$ samples, devoid of labels. Our model leverages both $X_L$ and $X_U$ in the SSL process, employing our proposed strategy of dynamic thresholding and re-sampling that's based on class-level confidence.

\subsection{Baseline: FixMatch}
\label{sec:fixmatch}
Consistency regularization is a well-established constraint in recent semi-supervised learning (SSL) algorithms ~\cite{berthelot2019mixmatch,berthelot2019remixmatch,tarvainen2017mean}. It ensures that the prediction results from various augmentations of the same instance remain coherent:
\begin{equation}
\sum\limits_{b=1}^{\mu B}{||p_m(y|\hat{\alpha}(u_b)) - p_m(y|\alpha(u_b))||}^2_2
\end{equation}

In the presented equation, several variables and parameters are defined as follows: $B$: refers to the batch size of the labeled data. $\mu$: denotes the ratio of labeled data to unlabeled data.  $\alpha$ and $\hat{\alpha}$ are functions signifying random data augmentation. $u_b$: represents the subset of unlabeled data that is drawn from the larger unlabeled dataset $X_U$. $p_m$: refers to the output probability as produced by our models.

Pseudo-labeling is a commonly utilized technique within SSL. At its core, this method leverages the model's predictions on unlabeled data to produce "pseudo-labels."  This approach sets a static threshold to exclude unlabeled data with low confidence, thus obtaining pseudo labels. The discrepancy between predictions and hard pseudo-labels is minimized using cross-entropy loss:
\begin{equation}
\begin{aligned}
\frac{1}{\mu B} \sum\limits_{b=1}^{\mu B} \mathbb{1}(max(p_m(y|\alpha(u_b)) \geq \tau)\ \cdot H(\hat{p}_m(y|\alpha(u_b),p_m(y|\alpha(u_b))
\end{aligned}
\end{equation}
In this context, $\hat{p}_m(y|\alpha(u_b) = \arg\max(p_m(y|\alpha(u_b))$ and the threshold is presented by $\tau$. $H(p,q)$ stands for the cross-entropy loss between $p$ and $q$. Typically, a high threshold value, denoted as $\tau$, is utilized to filter pseudo-labels that have low prediction confidence.

Recently, FixMatch~\cite{sohn2020fixmatch} has successfully integrated consistency regularization and pseudo-labeling, thereby achieving SOTA results on numerous tasks~\cite{wang20213dioumatch,zhao2020sess,chen2021multimodal}. FixMatch comprises a supervised loss $\ell_s$, and an unsupervised loss $\ell_u$. We define the supervised loss as:
\begin{equation}
\ell_s = \frac{1}{B} \sum\limits_{b=1}^{B}H(y_b,p_m(y|\alpha(x_b))
\end{equation}
Here, $y_b$ symbolizes the label of the labeled instance $x_b$. Regarding the unsupervised loss, $\ell_u$, FixMatch chooses predictions that are confident (exceeding the threshold) from the weakly augmented data to act as pseudo labels. Subsequently, a minimization of the cross-entropy loss is carried out using the prediction results derived from the data's strongly augmented perspectives, along with the associated pseudo label. The expression for the unsupervised loss is as follows:
\begin{equation}
\begin{aligned}
\ell_u = \frac{1}{\mu B} \sum\limits_{b=1}^{\mu B} \mathbb{1}(max(p_m(y|\alpha(u_b)) \geq \tau)\ \\ \cdot H(\hat{p}_m(y|\alpha(u_b),p_m(y|\mathcal{A}{(u_b)})
\end{aligned}
\end{equation}
In this context, $\mathcal{A}$ represents a function for strong data augmentation. Due to the straightforward nature and exceptional performance of FixMatch, it serves as a model for many contemporary SSL methods across various tasks. However, within FixMatch and its associated variations, a high constant value $\tau$ is often set for the threshold. While setting such a high threshold can improve the quality of pseudo-labels, it simultaneously diminishes the count of pseudo-labels contributing to the network's optimization, thereby leaving a substantial portion of the unlabeled data unused.

\subsection{Class-Level Confidence Based Dynamic Threshold}
\label{sec:threshold}



Drawing upon the concepts introduced in FlexMatch [1], we introduce a dynamic thresholding strategy that hinges on class-level confidence, thereby utilizing the learning status of each individual class. The prevalent approach in Semi-Supervised Learning (SSL) methodologies is to rely on instance-level prediction confidence for assessing the quality of each instance. In contrast, our approach seeks to tap into the class-level confidence. Our analysis reveals that the class-level confidence for each class, as it pertains to unlabeled data, can effectively embody the learning status of that class. The usage of this estimated learning status empowers us to adjust the threshold for each class dynamically, thereby augmenting the efficiency of SSL. We use prediction probability $argmax$ to obtain each class's unlabeled set:{$\{ C_{c} | u_b \in N_U , \arg\max(p_m(y|\alpha(u_b)))=c, b = 1, 2, ..., \mu B \}$. Then, we average the confidence of each unlabeled class to obtain class-level confidence.
\begin{equation}
P_{c} = \frac{1}{|C_{c}|}\sum_{j=0}^{|C_{c}|} \max(p_m(y|\alpha(u_j)), u_j\in C_{c}
\end{equation}
Subsequently, we use a non-linear function to obtain the learning status from class-level confidence. The complete procedure to derive the dynamic threshold is formulated as follows:
\begin{equation}
\label{eq2}
\tau_{e}(c) =
\begin{cases}
 1 - \tau,  \hspace{0.1cm} & if \hspace{0.1cm}  M(P_{c}) < 1 - \tau  \\
 \tau,  \hspace{0.1cm} & elif \hspace{0.1cm}  M(P_{c}) > \tau \\
 M(P_{c}), \hspace{0.1cm} & else  \\
\end{cases}
\end{equation}

In this context, $P_{c}$ symbolizes the class-level confidence for class $c$. $M(x) = \frac{x}{2-x}$ is a non-linear function serving as a mapping tool. $\tau_{e}(c)$ represents the dynamic threshold for class $c$ at the epoch $e$. Lastly, $\tau$ denotes the comprehensive dynamic threshold, which is obtained by evaluating the overall confidence.

\begin{equation}
\tau = exp{(-2 \times (P_{ave})^2)}
\end{equation}

Since unlabeled data lack labels, we leverage the network's maximum prediction ($argmax$) as the class for each unlabeled instance $u_i$, denoted as $c_{i}=\arg\max(p_m(y|\alpha(u_i)))$. $P_{ave}$ refers to the average value of all class-level confidences, $P_c$. Our method filters high-quality pseudo-labels using dynamic threshold, defined as:
\begin{equation}
\hat{y}{c}^{i} = \mathbb{1}\left[ p_i\geq \tau{e}(c_{i})\right]
\end{equation}
where $p_i$ is the maximum model output probability given the input $u_i$ augmented by $\alpha$, i.e., $p_i=\max(p_m(y|\alpha(u_i)))$. We decrease the threshold for classes with a lower learning status to enhance the number of pseudo-labels and subsequently improve learning. Since our approach modulates thresholds anchored on class-level confidence alone, it is applicable to any dataset type and boasts a high degree of generalizability. The unsupervised loss for our proposed dynamic threshold method is formulated as follows:

\begin{equation}
\ell_{u,e}=\frac{1}{\mu B} \sum\limits_{b=1}^{\mu B} H(\hat{y}_{c}^{i}, p_m(y|\mathcal{A}{(u_b)})
\end{equation} 

\subsection{Class-Level Confidence Based Data Re-sampling}
\label{sec:re-sampling}

Even though the number of pseudo-labels for classes experiencing lower learning status is increased by our proposed dynamic threshold method, it does not fully rectify the uneven learning state among classes. This inequity is caused by data imbalances and fluctuating levels of learning difficulty across different classes. For instance, consider the ModelNet40 dataset where the classes of airplane and bowl have 563 and 59 labeled data entries respectively. Even when the dynamic thresholding process excludes half of the airplane's pseudo-labels and incorporates all pseudo-labels from the bowl class, the quantity of chosen unlabeled data for the airplane class is at minimum four times that of the bowl class. Moreover, with its distinct shape, the airplane class inherently presents less learning difficulty compared to the bowl class. Therefore, despite the dynamic threshold, classes like the airplane maintain high learning status, potentially leading to overfitting due to lesser learning difficulty, the abundance of their data, or the combination of both. Such a situation can compromise the overall efficiency of the network.

To mitigate the issue, we turn to a re-sampling strategy, proven to be efficient in SSL settings where datasets are imbalanced. Most existing SSL methods address imbalance sample based on the quantity of data ~\cite{wei2021crest,he2021rethinking}. However, we observe that some minority classes could outperform majority classes as they have lower learning difficulty. A data quantity-based sampling approach can inadvertently bias the model towards classes with lower learning difficulty. To circumvent this, we introduce a re-sampling strategy based on class-level confidence. This method aims to boost the sampling probability of classes with a lower learning status, effectively considering both learning difficulty and data imbalance. We formulate each class's sampling probability as follows:
\begin{equation}
\centering
\label{eq2}
\begin{cases}
1 - W(e)\cdot P_c \cdot p_i  & ,  if \hspace{0.1cm} P_c > \tau \hspace{0.2cm} , \\
2 - W(e)\cdot P_c \cdot p_i  & ,  if \hspace{0.1cm} P_c \leq \tau \hspace{0.2cm} , \\
\end{cases}
\end{equation}
. Following previous work~\cite{yu2021playvirtual,ma2021adequacy}, we set the warm-function as $W(e)=exp{(-5 \times (1 - e/E_{max})^2)}$ This choice is deliberate to mitigate the risks associated with overly aggressive sampling. $p_i$  stands for the prediction confidence associated with the specific instance $i$. $c$ stands for the prediction class for instance $i$. The class-level confidence for class $c$ is indicated by $P_c$. $e$ refers to the current epoch, which indicates the present iteration of the training process. $E_{max}$ symbolizes the maximum epoch.

\subsection{Final Loss Function}
\label{sec:extend to general}
Our approach's core contributions are dynamic adjustment of the pseudo-label threshold and the re-sampling of data. Those strategies can be smoothly incorporated into other pseudo-label based approaches. The proposed method includes two distinct loss functions: the supervised loss, represented as $\ell_s$, for labeled data, and $\ell_{u,e}$ for data without labels. This relationship can be formulated as:
\begin{equation}
\ell = \ell_s + \ell_{u,e}
\end{equation}

\section{Experiments}

\begin{table*}[http]
    \centering

\adjustbox{width=330pt,height=390pt,keepaspectratio=true}
{

        \begin{tabular}{|p{10mm}|p{10mm}|p{10mm}|p{10mm}|p{10mm}|p{10mm}|p{10mm}|p{10mm}|p{10mm}|p{10mm}|p{10mm}|p{10mm}|}

	    \hline
        \multicolumn{2}{|c|}{\multirow{3}{*}{Dataset}}&\multicolumn{2}{c|}{\multirow{3}{*}{Method}} &\multicolumn{2}{c|}{2\% } &\multicolumn{2}{c|}{5\% } &\multicolumn{2}{c|}{10\% } &\multicolumn{2}{c|}{100\% }
        
        \\\cline{5-12}
        \multicolumn{2}{|c|}{} & \multicolumn{2}{c|}{} & Overall & Mean& Overall& Mean& Overall& Mean & Overall& Mean
        \\
        \multicolumn{2}{|c|}{} & \multicolumn{2}{c|}{} & Acc & Acc& Acc& Acc& Acc& Acc & Acc& Acc\\

        \hline
        \multicolumn{2}{|c|}{\multirow{7}{*}{ ModelNet40}} & \multicolumn{2}{c|}{Point Transformer~\cite{guo2021pct}} & 71.1 & 61.0 & 77.1& 69.2& 84.6& 77.2 & 93.2 &85.6
        \\\cline{3-12}
        \multicolumn{2}{|c|}{} & \multicolumn{2}{c|}{PL~\cite{lee2013pseudo}} & 69.7  & 59.6 & 78.3& 69.0& 85.1& 77.7 & 93.3 &85.9
        \\\cline{3-12}
        \multicolumn{2}{|c|}{} &
        \multicolumn{2}{c|}{Flex-PL~\cite{zhang2021flexmatch}} & 66.7 & 54.9 & 74.2& 62.3& 83.2& 70.3 & 93.3 &86.3
        \\\cline{3-12}
        \multicolumn{2}{|c|}{} &
        \multicolumn{2}{c|}{Confid-PL~\cite{chen2023class}} & 74.4 & 61.9 & 80.6& 73.5& 86.5 & 80.4 & 93.5 &86.6
        \\\cline{3-12}
        \multicolumn{2}{|c|}{} &
        \multicolumn{2}{c|}{DyConfid-PL(Ours)} & \textbf{77.8} & \textbf{64.2} & \textbf{83.1}& \textbf{76.9}& \textbf{88.3}& \textbf{82.1} & \textbf{93.5}&\textbf{86.9}
        \\\cline{3-12}
        \multicolumn{2}{|c|}{} & \multicolumn{2}{c|}{FixMatch~\cite{sohn2020fixmatch}(NeurIPS 2020)} & 70.8 & 62.7 & 78.9& 71.1& 85.5& 79.4 & 93.4 &86.6
        \\\cline{3-12}
        \multicolumn{2}{|c|}{} & \multicolumn{2}{c|}{Dash~\cite{xu2021dash}(ICML 2021)} & 71.5 & 63.0 & 79.7& 71.8& 85.9& 80.1 & 93.3 &86.4
        \\\cline{3-12}
        \multicolumn{2}{|c|}{} & \multicolumn{2}{c|}{FlexMatch~\cite{zhang2021flexmatch}(NeurIPS 2021)} & 70.1 & 61.2 & 80.5& 70.4& 86.2& 78.7  & 93.2 &86.1
        \\\cline{3-12}
        \multicolumn{2}{|c|}{} & \multicolumn{2}{c|}{Confid-Match~\cite{chen2023class}} & 73.8 & 64.1 & 82.1& 74.3& 87.8& 82.5 & 93.5 &86.8
        \\\cline{3-12}
        \multicolumn{2}{|c|}{} & \multicolumn{2}{c|}{DyConfid-Match(Ours)} & \textbf{76.5} & \textbf{66.7} & \textbf{83.8}& \textbf{76.4}& \textbf{88.8}& \textbf{84.6} & \textbf{93.7} &\textbf{87.3}
        \\
        \hline
	\end{tabular}
}
\caption{Results comparison with state-of-the-art 3D semi-supervised object classification methods on the ModelNet40 dataset.}
\label{tab:Classification-modelnet}
\end{table*}

\subsection{Datasets}
\textbf{Classification Datasets:} We evaluate our technique using two widely recognized datasets, ScanObjectNN ~\cite{uy-scanobjectnn-iccv19} and ModelNet40~\cite{wu20153d}. ModelNet40, a frequently employed benchmark, encompasses 12,311 Computer-Aided Design (CAD) models featuring mesh structures from 40 categories. The dataset is partitioned into 9,843 training and 2,468 testing models. On the other hand, ScanObjectNN provides a point cloud object dataset that is more reflective of real-world conditions. It incorporates 15,000 objects spanning 15 classes, all derived from 2,902 distinct real-world instances.

\textbf{Detection Datasets:} Following previous 3D SSL object detection methods ~\cite{wang20213dioumatch}, we validate our method with two prevalent detection benchmarks: SUN RGB-D~\cite{song2015sun} and ScanNet~\cite{dai2017scannet}. ScanNet ~\cite{dai2017scannet} is a dataset of indoor scenes containing 1,513 reconstructed meshes. Out of these, 1,201 serve as training samples and 312 as validation samples. Conversely, SUN RGB-D~\cite{song2015sun}, with over 10,000 indoor scenes, apportions 5,285 scenes for training and 5,050 for validation.

\subsection{Implementation Details}

\begin{table*}[t!]
    \centering

\adjustbox{width=330pt,height=390pt,keepaspectratio=true}
{

        \begin{tabular}{|p{15mm}|p{15mm}|p{15mm}|p{15mm}|p{15mm}|p{15mm}|p{15mm}|p{15mm}|p{15mm}|p{15mm}|}

	    \hline
        \multicolumn{2}{|c|}{\multirow{3}{*}{Dataset}}&\multicolumn{2}{c|}{\multirow{3}{*}{Method}} &\multicolumn{2}{c|}{1\% } &\multicolumn{2}{c|}{2\% } &\multicolumn{2}{c|}{5\%} 
        
        \\\cline{5-10}
        \multicolumn{2}{|c|}{} & \multicolumn{2}{c|}{} & Overall & Mean& Overall& Mean& Overall& Mean
        \\
        \multicolumn{2}{|c|}{} & \multicolumn{2}{c|}{} & Acc & Acc& Acc& Acc& Acc& Acc\\
        
        \hline
        \multicolumn{2}{|c|}{\multirow{7}{*}{ScanObjectNN}} & \multicolumn{2}{c|}{Point Transformer~\cite{guo2021pct}} & 32.1 & 26.1 & 44.7& 36.5& 56.6& 50.0  
        \\\cline{3-10}
        \multicolumn{2}{|c|}{} & \multicolumn{2}{c|}{PL~\cite{lee2013pseudo}} & 31.2 & 25.8 & 47.5& 38.6& 58.1& 51.5
        \\\cline{3-10}
        
        \multicolumn{2}{|c|}{} &
        \multicolumn{2}{c|}{Flex-PL~\cite{zhang2021flexmatch}} & 29.2 & 24.2 & 47.2& 37.6& 60.1& 51.9
        \\\cline{3-10}
        \multicolumn{2}{|c|}{} &
        \multicolumn{2}{c|}{Confid-PL~\cite{chen2023class}} & 34.2 & 28.9 & 50.2& 43.3& 64.8& 57.8
        \\\cline{3-10}
        \multicolumn{2}{|c|}{} &
        \multicolumn{2}{c|}{DyConfid-PL(Ours)} & \textbf{35.6} & \textbf{30.1} & \textbf{52.8}& \textbf{45.5}& \textbf{65.3}& \textbf{59.2}
        \\\cline{3-10}
        
        \multicolumn{2}{|c|}{} & \multicolumn{2}{c|}{FixMatch~\cite{sohn2020fixmatch}(NeurIPS 2020)} & 33.5 & 27.6 & 47.4& 39.9& 59.4& 52.4
        \\\cline{3-10}
        \multicolumn{2}{|c|}{} & \multicolumn{2}{c|}{Dash~\cite{xu2021dash}(ICML 2021)} & 35.1 & 29.3 & 50.3& 44.1& 62.8& 60.3
        \\\cline{3-10}
        \multicolumn{2}{|c|}{} & \multicolumn{2}{c|}{FlexMatch~\cite{zhang2021flexmatch}(NeurIPS 2021)} & 34.2 & 26.2 & 48.5& 39.7& 63.4& 57.2
        \\\cline{3-10}
        \multicolumn{2}{|c|}{} & \multicolumn{2}{c|}{Confid-Match~\cite{chen2023class}} & 38.2 & 32.7 & 57.0& 48.6& 69.4& 65.5
        \\\cline{3-10}
        \multicolumn{2}{|c|}{} & \multicolumn{2}{c|}{DyConfid-Match(Ours)} & \textbf{40.5} & \textbf{34.8} & \textbf{59.3}& \textbf{50.5}& \textbf{71.0}& \textbf{67.2}
        \\
        \hline

	\end{tabular}
}\caption{Results comparison with state-of-the-art 3D semi-supervised object classification methods on the ScanObjectNN dataset.
}
\label{tab:Classification-objectnn}
\end{table*}
\textbf{3D Object Classification:} We leverage the Stochastic Gradient Descent (SGD) optimizer and set the learning rate to 0.01 in both ModelNet40 and ScanObjectNN datasets. We utilize the CosineAnnealingLR decay to schedule the learning rate and set the minimum rate as 0.0001. Models are trained over a span of 500 epochs. We leverage
 rotation and arbitrary scaling as weak augmentation, and rotation, arbitrary scaling, translation, jittering, and random scaling as strong augmentation. The entire batch comprises 240 samples, of which 48 are labeled data. The remaining 192 samples are unlabeled. We equally set the weight for both supervised and unsupervised as 1. A threshold is established at $\tau = 0.8$. The data loader undergoes updates every 50 epochs in line with the re-sampling strategy. For the SSL classification task, PointTransformer~\cite{guo2021pct} is utilized as the backbone.

\textbf{3D Object Detection:} We leverage the ground-breaking 3DIoUMatch approach in our method, retaining similar settings. However, rather than employing the pre-trained model from VoteNet and PVRCNN~\cite{shi2020pv}, as 3DIoUMatch does, we introduce our proposed re-sampling strategy during the pre-training phase. The weights obtained during pre-training are subsequently used to initialize models. For scenes with multiple objects, we select the lowest confidence object. Similar to the classification task, the data loader undergoes updates every 50 epochs in sync with the re-sampling strategy. In order to make the comparison fair, we leverage the same pre-processing strategies and labels as previous works~\cite{qi2019deep,wang20213dioumatch}. Furthermore, the mean average precision (mAP) under intersection over union (IoU) thresholds of 0.25 and 0.5 are employed as the evaluation metric. The threshold is similarly set to $\tau = 0.8$.

\begin{table*}[t!]
\setlength{\belowcaptionskip}{-2pt}
    \centering

\adjustbox{width=350pt,height=950pt,keepaspectratio=true}
{

        \begin{tabular}{|p{15mm}|p{15mm}|p{15mm}|p{15mm}|p{15mm}|p{15mm}|p{15mm}|p{15mm}|p{15mm}|p{15mm}|p{15mm}|p{15mm}|p{15mm}|p{15mm}|}

	    \hline
        \multicolumn{2}{|c|}{\multirow{3}{*}{Dataset}}&\multicolumn{2}{c|}{\multirow{3}{*}{Method}} &\multicolumn{2}{c|}{1\% } &\multicolumn{2}{c|}{2\% } &\multicolumn{2}{c|}{5\% } 
        &\multicolumn{2}{c|}{20\% } 
        &\multicolumn{2}{c|}{100\% } 
        \\\cline{5-14}
        \multicolumn{2}{|c|}{} & \multicolumn{2}{c|}{} & mAP & mAP& mAP& mAP& mAP& mAP & mAP& mAP & mAP& mAP
        \\
        \multicolumn{2}{|c|}{} & \multicolumn{2}{c|}{} & $@$ 0.25 & $@$0.50& $@$0.25& $@$0.50& $@$0.25& $@$0.50 &$@$0.25& $@$0.50 & $@$0.25& $@$0.50\\
        \hline
        \multicolumn{2}{|c|}{\multirow{4}{*}{SUN  RGB-D}} & \multicolumn{2}{c|}{VoteNet~\cite{qi2019deep}(ICCV 2019)} & 16.7$\pm$1.2 & 3.9$\pm$0.9 & 21.8$\pm$1.6& 5.1$\pm$0.
        8& 33.9$\pm$1.9& 13.1$\pm$1.7 & 46.9$\pm$1.9 & 27.5$\pm$1.2& 57.8 &36.0 
        \\\cline{3-14}
        \multicolumn{2}{|c|}{} & \multicolumn{2}{c|}{SESS~\cite{zhao2020sess}(CVPR 2020)} & 19.9$\pm$1.6 & 6.3$\pm$1.2 &23.3$\pm$1.1 & 7.9$\pm$0.8& 36.1$\pm$1.1& 16.9$\pm$0.9 & 47.1$\pm$0.7 & 24.5$\pm$1.2 & 60.5 & 38.1
        \\\cline{3-14}
        \multicolumn{2}{|c|}{} & \multicolumn{2}{c|}{3DIoUMatch ~\cite{wang20213dioumatch} (CVPR 2021)} 
        & 25.6$\pm$0.6 
        & 9.4$\pm$0.7 
        & 26.8$\pm$0.7
        & 10.6$\pm$0.5
        & 39.7$\pm$0.9
        & 20.6$\pm$0.7
        &49.7$\pm$0.4 &30.9$\pm$0.2 &61.5 & 41.3
        \\\cline{3-14}
        \multicolumn{2}{|c|}{} & \multicolumn{2}{c|}{Confid-3DIoUMatch~\cite{chen2023class}} & 27.8$\pm$0.8 & 11.3$\pm$0.6 & 32.7$\pm$0.3 & 13.5$\pm$0.4& 43.1$\pm$0.6& 24.2$\pm$0.5
        &50.9$\pm$0.7 &31.5$\pm$0.8 &62.1 & 41.9
        \\\cline{3-14}
        \multicolumn{2}{|c|}{} & \multicolumn{2}{c|}{DyConfid-3DIoUMatch(Ours)} & \textbf{29.9$\pm$0.8} & \textbf{13.5$\pm$0.5} & \textbf{34.8$\pm$0.6}& \textbf{15.7$\pm$0.5}& \textbf{45.9$\pm$0.7}& \textbf{26.3$\pm$0.5}
        &\textbf{51.8$\pm$0.6} &\textbf{32.7$\pm$0.6} &\textbf{63.0} & \textbf{42.7}
        \\
        \hline

	\end{tabular}
}

 \caption{ 
 Results comparison with state-of-the-art 3D semi-supervised object detection methods on the SUN RGB-D
dataset.}
\label{tab:Detection-sun}
\end{table*}

\subsection{Semi-Supervised 3D Object Classification Results}
By comparing our method with a range of SSL approaches including Pseudo-Labeling (PL)~\cite{lee2013pseudo}, FixMatch~\cite{sohn2020fixmatch}, Dash~\cite{xu2021dash}, and FlexMatch~\cite{zhang2021flexmatch}, we evaluated the efficacy and potential of the proposed method. All comparisons are made under identical settings on ModelNet40~\cite{wu20153d} and ScanObjectNN~\cite{uy-scanobjectnn-iccv19}. FlexMatch~\cite{zhang2021flexmatch}, currently regarded as the leading-edge method, was developed to overcome the fixed-threshold drawback of FixMatch via dynamic threshold adjustment using curriculum pseudo labeling. To ensure fairness, all the techniques are applied using the same backbone, data augmentation, and hyperparameters.

Our method is integrated with both FixMatch and Pseudo-labeling for a comprehensive comparison against leading strategies. We present the results under varying proportions of labeled data. Evaluation of performance is done via two metrics: overall accuracy and mean class accuracy, which calculates the average accuracy across all classes. From the results in Table~\ref{tab:Classification-modelnet} and Table~\ref{tab:Classification-objectnn}, it's clear that FlexMatch, although the current front-runner in the 3D SSL classification task, shows marginal improvement or even a \textbf{decline} in performance when combined with FixMatch and Pseudo-Labeling given scarce labeled data. This outcome can be attributed to FlexMatch's design, which is better suited to class-balanced datasets, while 3D datasets inherently possess data imbalance. As demonstrated in Table~\ref{tab:Classification-modelnet} and Table~\ref{tab:Classification-objectnn}, our model outperforms all other cutting-edge methods on both the ModelNet40 and ScanObjectNN datasets. The most substantial improvements can be observed in mean class accuracy, highlighting our model's proficiency in equalizing the learning status across classes.

\begin{table*}[]
\setlength{\belowcaptionskip}{-10pt}
    \centering

\adjustbox{width=350pt,height=950pt,keepaspectratio=true}
{

\begin{tabular}{|p{15mm}|p{15mm}|p{15mm}|p{15mm}|p{15mm}|p{15mm}|p{15mm}|p{15mm}|p{15mm}|p{15mm}|p{15mm}|p{15mm}|p{15mm}|p{15mm}|}

	    \hline
        \multicolumn{2}{|c|}{\multirow{3}{*}{Dataset}}&\multicolumn{2}{c|}{\multirow{3}{*}{Method}} &\multicolumn{2}{c|}{1\% } &\multicolumn{2}{c|}{2\% } &\multicolumn{2}{c|}{5\% } &\multicolumn{2}{c|}{20\% } &\multicolumn{2}{c|}{100\% } 
        \\\cline{5-14}
        \multicolumn{2}{|c|}{} & \multicolumn{2}{c|}{} & mAP & mAP& mAP& mAP& mAP& mAP & mAP& mAP & mAP& mAP
        \\
        \multicolumn{2}{|c|}{} & \multicolumn{2}{c|}{} & $@$ 0.25 & $@$0.50& $@$0.25& $@$0.50& $@$0.25& $@$0.50 & $@$0.25& $@$0.50 & $@$0.25& $@$0.50\\
        \hline
        \multicolumn{2}{|c|}{\multirow{4}{*}{ScanNet}} & \multicolumn{2}{c|}{VoteNet~\cite{qi2019deep}(ICCV 2019)} & 8.9$\pm$1.1 & 1.5$\pm$0.5 & 16.9$\pm$1.3& 4.7$\pm$0.8& 31.2$\pm$1.1& 14.7$\pm$0.7 & 46.9$\pm$1.9 & 27.5$\pm$1.2 &57.8 & 36.0
        \\\cline{3-14}
        \multicolumn{2}{|c|}{} & \multicolumn{2}{c|}{SESS~\cite{zhao2020sess}(CVPR 2020)} &11.3$\pm$1.6&2.7$\pm$0.6& 21.1$\pm$1.5&8.4$\pm$1.1&35.5$\pm$2.0&17.2$\pm$0.9
        & 49.6$\pm$1.1 &29.0$\pm$1.0 &61.3 &39.0
        \\\cline{3-14}
        \multicolumn{2}{|c|}{} & \multicolumn{2}{c|}{3DIoUMatch ~\cite{wang20213dioumatch} (CVPR 2021)} & 14.6$\pm$1.4&3.9$\pm$0.5 & 24.5$\pm$1.9& 11.2$\pm$1.4&40.4$\pm$0.8&21.0$\pm$0.6
        &52.8$\pm$1.2& 35.2$\pm$1.1& 62.9 &42.1
        \\\cline{3-14}
        \multicolumn{2}{|c|}{} & \multicolumn{2}{c|}{Confid-3DIoUMatch~\cite{chen2023class}} & 19.0$\pm$0.4 & 6.4$\pm$0.4 & 29.5$\pm$1.5& 15.2$\pm$0.6& 43.6$\pm$0.5& 24.3$\pm$0.4         &53.6$\pm$1.5& 36.4$\pm$0.8& 63.2 &42.7
        \\\cline{3-14}
        \multicolumn{2}{|c|}{} & \multicolumn{2}{c|}{DyConfid-3DIoUMatch(Ours)} & \textbf{22.3$\pm$0.7} & \textbf{8.3$\pm$0.5} & \textbf{32.2$\pm$1.4}& \textbf{18.3$\pm$0.8}& \textbf{46.2$\pm$0.8}& \textbf{26.8$\pm$0.7} &
        \textbf{53.7$\pm$1.3}& \textbf{37.3$\pm$1.4}& \textbf{63.7} &\textbf{43.1}
        \\
        \hline

	\end{tabular}
}
 \caption{
   Results comparison with state-of-the-art 3D semi-supervised object detection methods on the ScanNet
dataset.}
\label{tab:Detection-scan}
\end{table*}

\subsection{Semi-Supervised 3D Object Detection Results}

We additionally conduct a comparison between our approach and the SOTA 3D SSL object detection methods. We enhanced the 3DIoUMatch~\cite{wang20213dioumatch} approach by integrating our proposed methods with 3DIoUMatch. This refined model was then evaluated against SOTA methods such as VoteNet~\cite{qi2019deep}, SESS~\cite{zhao2020sess}, and 3DIoUMatch~\cite{wang20213dioumatch} on two indoor datasets: SUN RGB-D and ScanNet. We adhered to established norms and calculated the mean average precision (mAP) under two separate thresholds of $0.25$ and $0.5$. We also evaluate our work in the outdoor detection KITTI~\cite{geiger2013vision} dataset as 3DIoUMatch.

The outcomes presented in Table~\ref{tab:Detection-sun}, Table~\ref{tab:Detection-scan}, and Table~\ref{tab:kitti_main} emphasize that the proposed method outperforms all other SOTA methods across the SUN RGB-D, ScanNet, and KITTI datasets across various scenarios. Notably, our method generates the most remarkable improvement when considering the 2\% labeled scenario, where our approach outperforms 3DIoUMatch by a remarkable \textbf{7} and \textbf{6.7} in terms of mAP@0.25 on the ScanNet and SUN RGB-D datasets, respectively. These findings validate the efficacy of integrating our innovative class-level confidence-based dynamic threshold and learning status balancing re-sampling strategy into other semi-supervised techniques, leading to enhanced performance.

\subsection{Ablation Study for Dynamic Thresholding and Re-sampling}

The proposed approach comprises two primary elements: a dynamic thresholding based on confidence levels and a dynamic re-sampling mechanism. To understand the significance of each element, we carried out ablation studies, evaluating different component mixtures for both the SSL 3D object classification and detection tasks. For the classification task, we built upon FixMatch~\cite{sohn2020fixmatch}, while using 3DIoUMatch~\cite{wang20213dioumatch} as the foundation for the detection task. Results pertaining to the classification tasks can be found in Table~\ref{tab:detection_abl}.

\textbf{Comprehensive
Dynamic Thresholding:} 
In contrast to the fixed maximum threshold like 0.9 utilized in the previous conference version, we introduce a comprehensive
dynamic thresholding for training. This innovative approach removes the necessity for a manually set threshold, enabling the network to autonomously determine the most appropriate threshold during the training process. Our dynamic thresholding strategy adapts in real time, optimizing the threshold based on the specific requirements of the data and the learning task at hand. Implementing this dynamic maximum threshold approach has led to notable enhancements in performance across a wide array of tasks and scenarios. In several instances, the improvements have been particularly significant, underscoring the robustness and versatility of our method. The observed gains are not just marginal but substantial, indicating a profound impact on the overall learning efficiency and accuracy. These promising results validate the efficacy of our dynamic thresholding approach. 

\begin{table*}[t]
\setlength{\belowcaptionskip}{-10pt}
    \centering

\adjustbox{width=320pt,height=700pt,keepaspectratio=true}
{
    \begin{tabular}{|c|c|c|c|c|c|c|c|c|c|}
    \hline
    & \multicolumn{3}{c|}{1\%} & \multicolumn{3}{c|}{2\%} & \multicolumn{3}{c|}{100\%}\\ 
      \cline{2-10} \multirow{-2}{*}{} & Car & Ped. & Cyc. & Car & Ped. & Cyc. & Car & Ped. & Cyc.  \\ \hline
    PVR ~\cite{shi2020pv} & 73.5 & 28.7 & 28.4 & 76.6 & 40.8 & 45.5 & 83.0 & 57.9 & 73.1 \\ \cline{1-10} 
    3DIoUMatch ~\cite{wang20213dioumatch} & 76.0 & 31.7 & 36.4 & 78.7 & 48.2 & 56.2 & 84.8 & 60.2 & 74.9 \\ \cline{1-10} 

    DyConfid-3DIoUMatch  & \textbf{79.6} & \textbf{34.8} & \textbf{42.3} & \textbf{81.9} & \textbf{54.2} & \textbf{62.8} & \textbf{88.1} & \textbf{62.6} & \textbf{77.3} \\ \cline{1-10} 
    \end{tabular}
    }
\vspace{-1mm}
    \caption{Results comparison with state-of-the-art 3D semi-supervised object detection methods on the KITTI dataset.}
    \label{tab:kitti_main}
    
\end{table*}

\vspace{5mm}

\begin{table*}[t!]

\centering
\adjustbox{width=320pt,height=500pt,keepaspectratio=true}
{
	\begin{tabular}{|c|c|c|c|c|c|c|}

	    \hline
	    \multirowcell{2}{Comprehensive\\Dynamic Threshold} &
	    \multirowcell{2}{Confidence \\Re-sample} & \multirowcell{2}{ Class-level \\Dynamic Threshold} & 
	    \multicolumn{2}{c|}{ModelNet40 10\%} &
	    \multicolumn{2}{c|}{ObjectNN 5\%} \\ \cline{4-7}
	    & & & mAP @0.25 & mAP @0.5 & mAP @0.25 & mAP @0.5
        \\

        \hline
        & &  & 85.5&79.4& 59.4 & 52.4  \\
        \hline
        \checkmark& &  &86.5 &80.9  & 63.6 & 59.8  \\
        \hline
        \checkmark&\checkmark&  &  87.3 & 82.8 & 66.7 &  62.1  \\
        \hline
        \checkmark&  & \checkmark& 87.6 & 83.3 & 68.7 &  65.2  \\
        \hline
        \checkmark& \checkmark& \checkmark & \textbf{88.8} & \textbf{84.6} & \textbf{71.0}&\textbf{67.2}      \\
        \hline

	\end{tabular}
}
 \caption{Components effect ablation  study for 3D SSL object classification.}
\label{tab:detection_abl}
\end{table*}

\textbf{Dynamic Re-sampling:} In the context of baseline models used for both classification and detection, there is typically no re-sampling strategy applied, meaning that each data sample's probability is treated identically. However, for the SSL 3D classification task, the implementation of our specially designed dynamic re-sampling method generally leads to performance enhancements, with the most significant improvements observed on the scanObjectNN dataset. In 3D SSL detection tasks, leveraging this re-sampling approach during the pre-training phase, specifically for re-sampling labeled data, significantly enhances performance. The benefits are even more improved when re-sampling is applied during the semi-supervised training phase as well. Benchmark results for both SSL 3D classification and detection solidly confirm the benefits of our suggested dynamic re-sampling method.

\textbf{Class-level Dynamic Thresholding:} 
Unlike previous methods that use a uniform threshold for all classes, our approach considers the learning difficulty and data imbalance to derive a dynamic threshold adaptive for each class. Incorporating our dynamic thresholding strategy, based on class levels, enhances performance across both tasks under various scenarios, with some showing substantial improvement. These findings validate the efficacy of our class-level dynamic thresholding method, showcasing its potential for smooth integration with other SSL methods.

\subsection{Ablation Study on Other Design Choices}

To gain a more comprehensive understanding of our approach, we conducted detailed ablation studies. These studies were specifically aimed at understanding the roles and impacts of the threshold's maximum limits, the mapping functions utilized, and constants selection for the mapping function. We present the ablation study results for two distinct scenarios: firstly, SSL 3D classification tasks involving a dataset with a mere $10$ percent labeled data from ModelNet40; and secondly, SSL 3D object detection tasks that incorporated a scant $5$ percent labeled data from the ScanNet dataset.

\begin{table}[t!]
    \begin{subtable}[h]{0.45\textwidth}
         \centering
        \begin{tabular}{c|cc}
            \multirowcell{2}{Mapping\\Function}  & Overall & Overall \\  {}& @Acc & @Acc\\\hline
            Linear &$87.9$ & $83.0$\\
            Concave & $\textbf{88.8}$& $\textbf{84.6}$\\
            Exp & $87.2$& $83.8$\\
        \end{tabular}
        \captionsetup{width=0.9\textwidth}
        \caption{ 3D classification mapping function.}
        \label{tab:classification-mapping-abs}
    \end{subtable}
    \hspace{3mm}
    \begin{subtable}[h]{0.45\textwidth}
         \centering
        \begin{tabular}{c|cc}
            \multirowcell{2}{Mapping\\Function}  & mAP & mAP \\  {}& @0.25 & @0.5\\\hline
            Linear &$44.6$ & $24.3$\\
            Concave & $\textbf{45.2}$& $\textbf{25.8}$\\
            Exp & $44.0$& $24.9$\\
        \end{tabular}
        \captionsetup{width=.85\linewidth}
        \caption{3D detection mapping function .}
        \label{tab:detection-mapping-abs}
    \end{subtable}


    \caption{The ablation study for the effectiveness of the upper limit threshold and mapping function. }
    \label{tab:array}
\end{table}

\begin{table}[t!]
    \begin{subtable}[h]{0.45\textwidth}
         \centering
        \begin{tabular}{c|cc}
            \multirowcell{2}{Constants\\Selection}  & Overall & Overall \\  {}& @Acc & @Acc\\\hline
            1 &$85.8$ & $81.9$\\
            2 & $\textbf{88.8}$& $\textbf{84.6}$\\
            3 & $86.8$& $83.0$\\
        \end{tabular}
        \captionsetup{width=0.9\textwidth}
        \caption{ 3D classification constants selection.}
        \label{tab:constants-sel-abs-cls}
    \end{subtable}
    \hspace{3mm}
    \begin{subtable}[h]{0.45\textwidth}
         \centering
        \begin{tabular}{c|cc}
            \multirowcell{2}{Constants\\Selection}  & mAP & mAP \\  {}& @0.25 & @0.5\\\hline
            1 &$42.7$ & $22.6$\\
            2 & $\textbf{45.2}$& $\textbf{25.8}$\\
            3 & $43.8$& $23.9$\\
        \end{tabular}
        \captionsetup{width=.85\linewidth}
        \caption{3D detection constants selection .}
        \label{tab:constants-sel-abs-det}
    \end{subtable}


    \caption{ The ablation study for the effectiveness of the constants selection for the mapping function. }
    \label{tab:constants-sel}
\end{table}

\textbf{Learning Status Mapping Function:}

To delve even deeper into the impact of the learning status mapping function, we subjected three distinct mapping functions to testing: (1) exponential: $M(x_c) = \exp{(-5 \times (1 - P_c)^2)}$, (2) linear: $M(x_c) = P_c$, and (3) concave: $M(x_c) = P_c/(2-P_c)$, where $P_c$ signifies the class-level confidence for class $c$. The comprehensive results pertaining to both the SSL 3D classification and detection tasks are available in Table~\ref{tab:classification-mapping-abs} and Table~\ref{tab:detection-mapping-abs}. Our investigations distinctly underscored that the concave function outperformed the others, emerging as the most effective for both classification and detection tasks. Moreover, it's worth noting that the performance of all three mapping functions displayed relatively comparable outcomes across both tasks. These findings provide a nuanced understanding of the intricate relationships between these mapping functions and task performance, further strengthening the overall insight into our methodology.
\begin{table}[t!]
    \begin{subtable}[h]{0.45\textwidth}
         \centering
        \begin{tabular}{c|cc}
            \multirowcell{2}{Constants\\Selection}  & Overall & Overall \\  {}& @Acc & @Acc\\\hline
            1 &$87.2$ & $83.5$\\
            2 & $\textbf{88.8}$& $\textbf{84.6}$\\
            3 & $88.0$& $83.9$\\
        \end{tabular}
        \captionsetup{width=0.9\textwidth}
        \caption{ 3D classification constants selection.}
        \label{tab:constants-sel-abs-cls}
    \end{subtable}
    \hspace{3mm}
    \begin{subtable}[h]{0.45\textwidth}
         \centering
        \begin{tabular}{c|cc}
            \multirowcell{2}{Constants\\Selection}  & mAP & mAP \\  {}& @0.25 & @0.5\\\hline
            1 &$44.1$ & $24.3$\\
            2 & $\textbf{45.2}$& $\textbf{25.8}$\\
            3 & $44.6$& $25.1$\\
        \end{tabular}
        \captionsetup{width=.85\linewidth}
        \caption{3D detection constants selection .}
        \label{tab:constants-sel-abs-det}
    \end{subtable}


    \caption{ The ablation study for the effectiveness of the constants selection for the comprehensive dynamic threshold. }
    \label{tab:constants-sel-com}
\end{table}

\textbf{Constants Selection for the Mapping Function:}

To demonstrate the impact of constant selection for the concave mapping function, we conducted an ablation study on both classification and detection tasks. We chose a constant value of 2 because we wanted the threshold to approach 1 as the confidence neared 1, effectively filtering out low-confidence instances. As illustrated in Table \ref{tab:constants-sel}, setting the constant to 2 achieved the best results. When the constant was set to 1, the dynamic threshold value approached 1 as class-level confidence neared 0.5, resulting in most unlabeled data being unused. Conversely, setting the constant to 3 caused the dynamic threshold value to remain around 0.5 even when class-level confidence was close to 1, leading to insufficient filtering of low-quality predictions and consequently harming network performance.

\textbf{Constants Selection for the Comprehensive Dynamic Threshold:}

To demonstrate the impact of constant selection for the comprehensive dynamic threshold, we conducted an ablation study on both classification and detection tasks. As illustrated in Table \ref{tab:constants-sel-com}, setting the constant to 2 achieved the best results across all evaluated metrics. Moreover, our method has shown robustness to variations in the selection of the constant for the comprehensive dynamic threshold. This robustness ensures that the method remains effective and efficient without the need for precise tuning, making it more practical for real-world applications.

\begin{figure}[t!]
\centering
\includegraphics[width = 0.6\textwidth]{
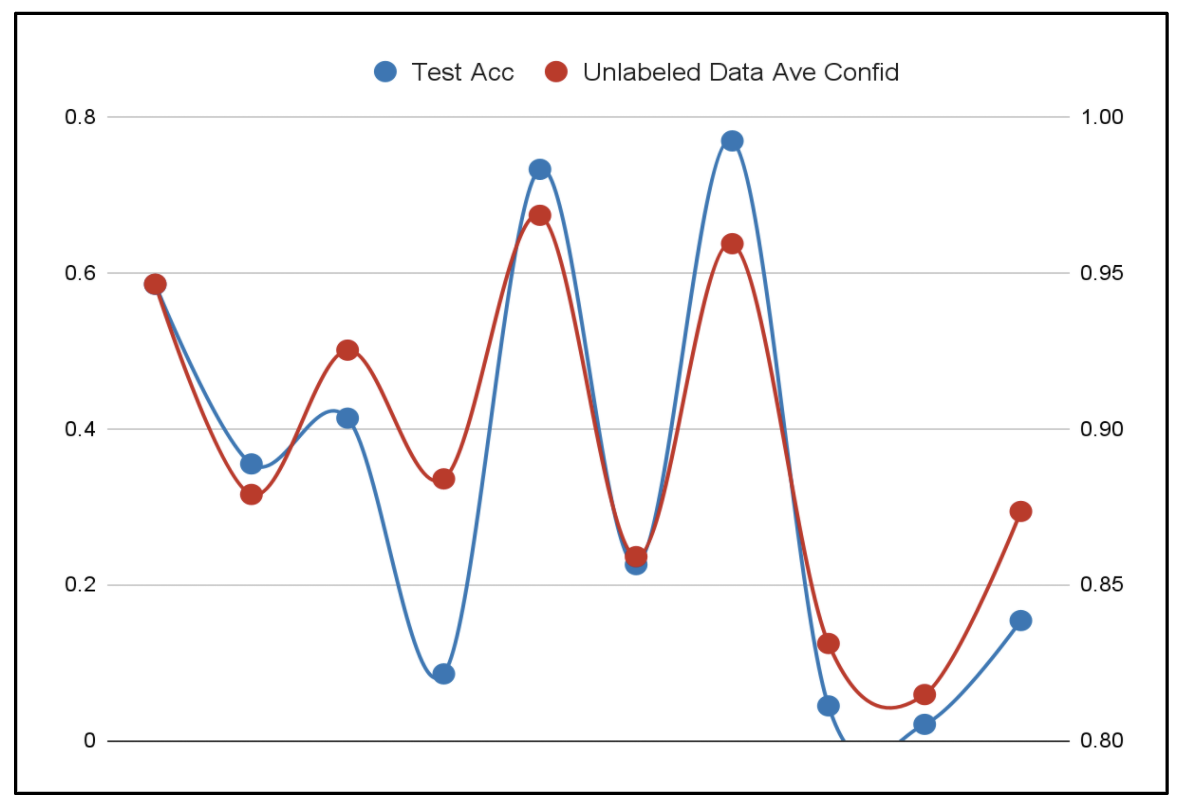}
\caption{Analysis of results for 3DIoUMatch in SUN-RGBD Dataset. The result demonstrates that the class-level confidence for unlabeled data exhibits a strong correlation with the class test accuracy in the 3D SSL detection task.} 
\label{fig:det-correlation}
\end{figure}

\subsection{Analysis of 3D SSL Object Detection}
We emphasize the connection between class-level confidence and the subsequent test accuracy in the context of 3D SSL classification. Our contention is that this class-level confidence potentially serves as an indicator of ongoing learning progress. In order to further substantiate the broad applicability of this observed relationship, we extend our investigation to include a 3D SSL object detection test utilizing the 3DIoUMatch approach, conducted on the SUN-RGBD dataset. For this detection test, we deliberately employed a meager 5 percent sample of the complete dataset. This correlation is provided in Fig~\ref{fig:det-correlation}. The compelling results of this study highlight a robust correlation between class-level confidence for unlabeled data and test accuracy in the context of detection tasks. This outcome serves to reinforce the universality of the hypothesis we've put forth, adding a new layer of confirmation to its significance across diverse scenarios.


\begin{figure*}[t!]
    \centering
    \begin{subfigure}[b]{0.45\textwidth}  
        \centering 
        \includegraphics[width=\textwidth]{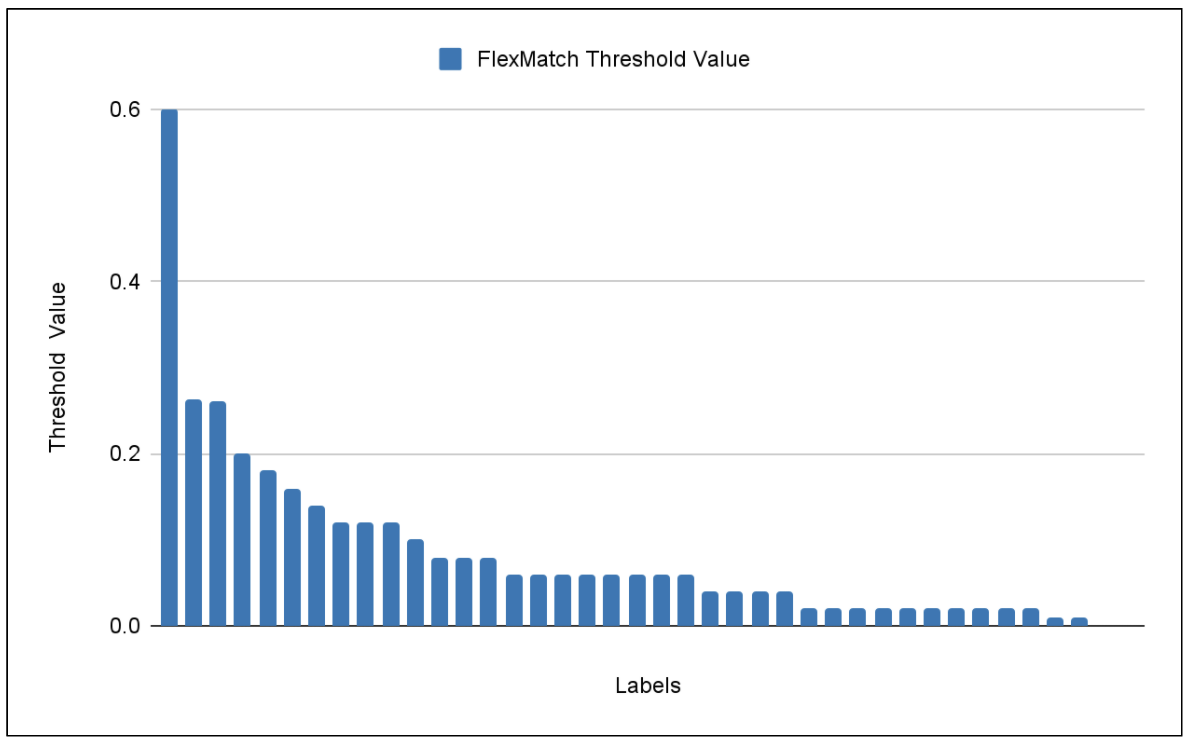}
        \caption[]%
        {{\small}}    
        \label{fig:flex-threshold}
    \end{subfigure}
    \hfill
    \begin{subfigure}[b]{0.45\textwidth}
        \centering
        \includegraphics[width=\textwidth]{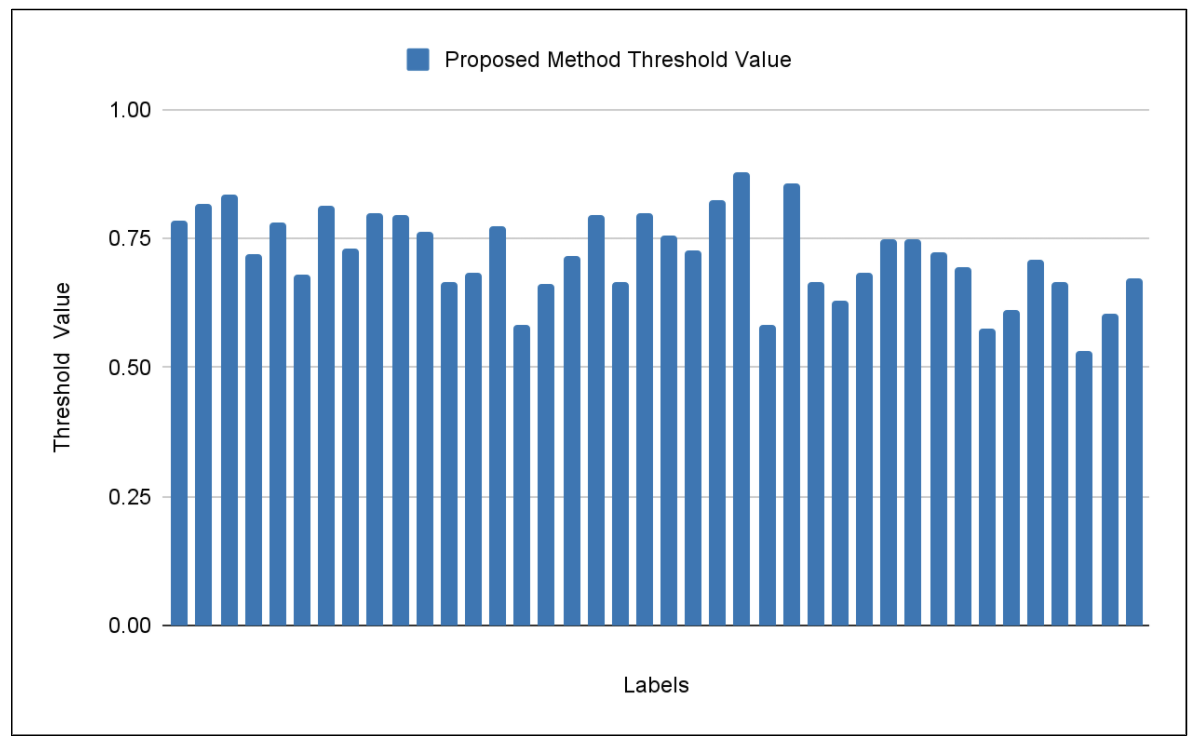}
        \caption[]%
        {{\small }}    
        \label{fig:our-threshold}
    \end{subfigure}
    \caption[ ]
    {\small (a) Thresholds for each class in the last epoch of our method, applied to the ModelNet40 dataset with $10\%$ labeled data. (b) Each class's threshold of the FlexMatch in the last epoch. FlexMatch results in long-tail thresholds, introducing significant noise into pseudo labels, thereby diminishing performance. In contrast, the thresholds in our method are well-proportioned and balanced, enhancing the effective use of unlabeled data.} 
    \label{fig:flexmatch-res}
\end{figure*}

\begin{figure*}[t!]
    \centering
    \begin{subfigure}[b]{0.3\textwidth}
        \centering
\includegraphics[width=\textwidth]{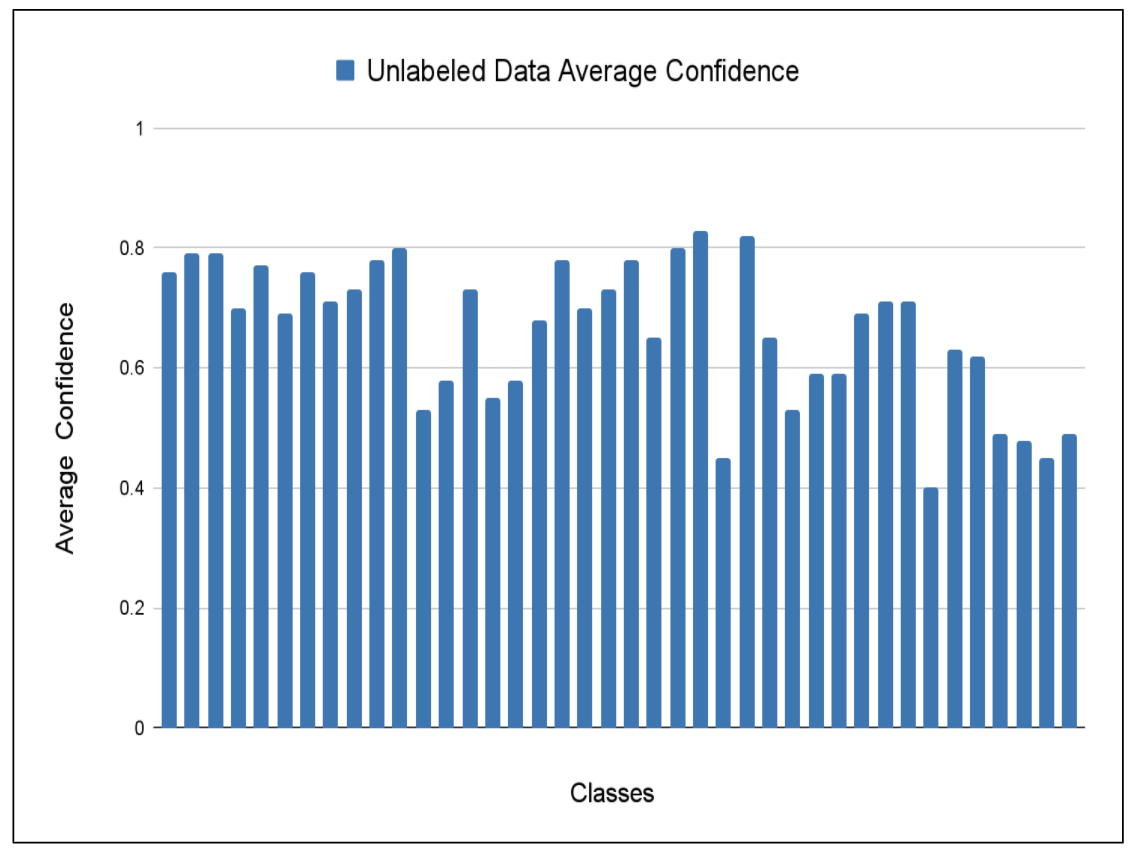}
        \caption[]%
        {{\small  The average confidence of FixMatch}}    
        \label{fig:modelnet_fixmatch_aveconfid}
    \end{subfigure}
    \begin{subfigure}[b]{0.3\textwidth}   
        \centering 
\includegraphics[width=\textwidth]{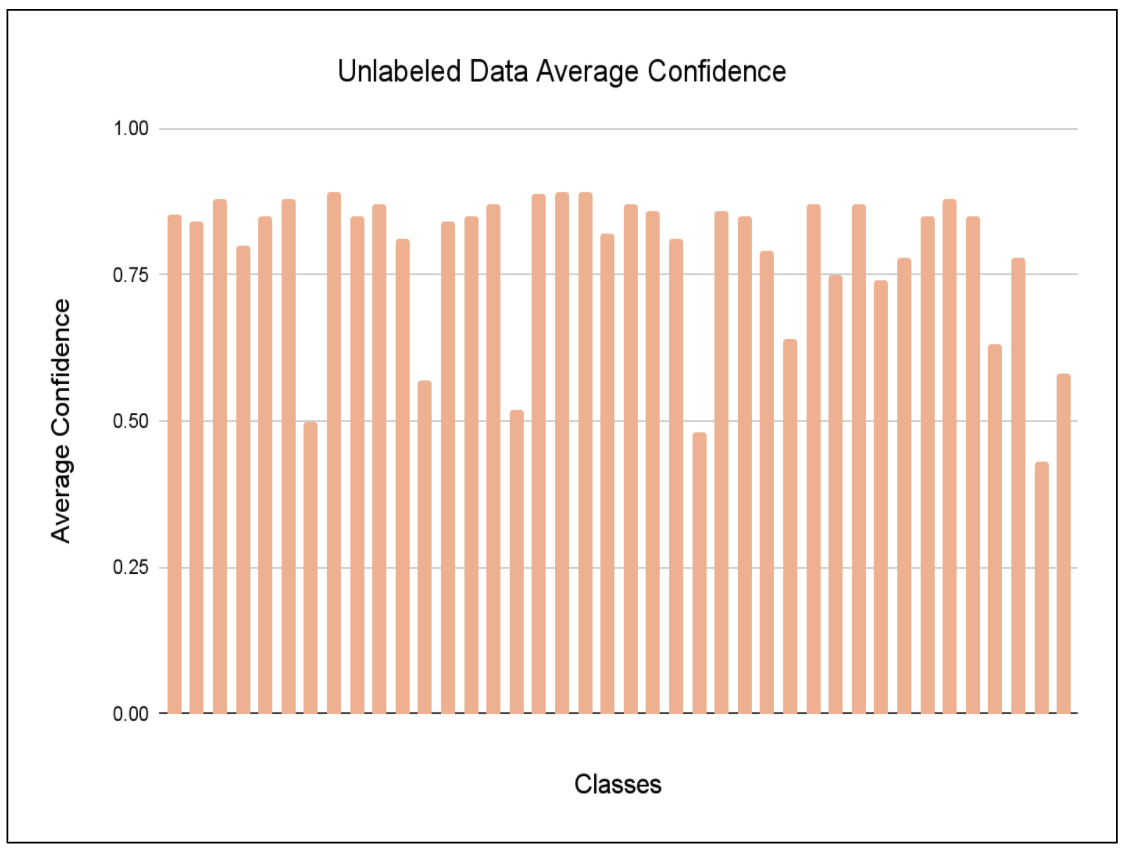}
        \caption[]%
        {{\small The average confidence of FlexMatch.}}    
        \label{fig:modelnet_flexmatch_aveconfid}
    \end{subfigure}
    \begin{subfigure}[b]{0.3\textwidth}   
        \centering 
\includegraphics[width=\textwidth]{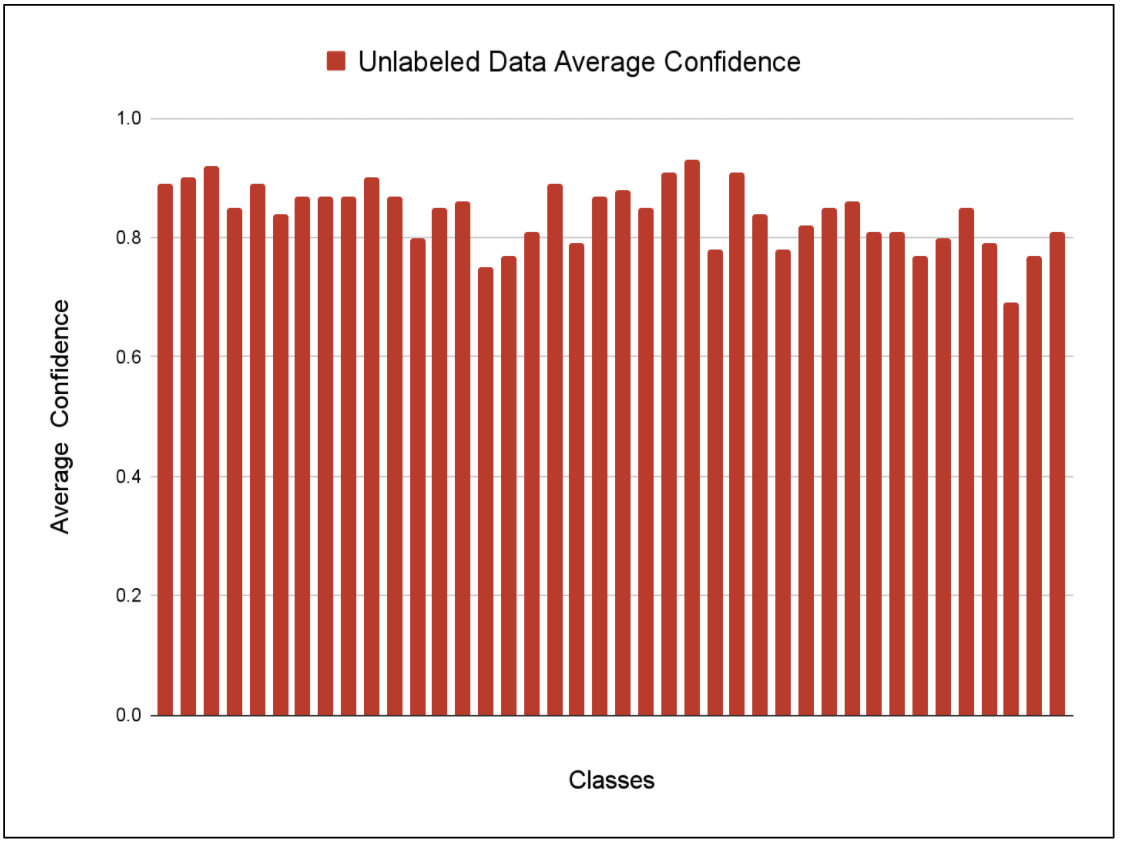}
        \caption[]%
        {{\small The average confidence of our method.}}    
        \label{fig:modelnet_our_aveconfid}
    \end{subfigure}
    \vspace{-5pt}

    \caption[ The average and standard deviation of critical parameters ]
    {\small Class-level confidence comparison in unlabeled data between FixMatch and our method, as applied to the ModelNet40 datasets and trained with $10\%$ labeled data, is presented. The findings reveal that our approach not only enhances the learning status for each class but also contributes to a more balanced and harmonized learning condition across classes.} 
    \label{fig:confidence comparison}
    \vspace{-10pt}
\end{figure*}

\subsection{Comparison with Other Class Imbalanced Resampling Method}

Most current methods for handling class imbalance in self-supervised learning (SSL) rely on data numbers in each class for resampling. However, we observe that some minority classes outperform the majority ones due to their lower learning difficulty. Resampling based solely on data numbers introduces a bias towards these easily learned classes. In contrast, our method prioritizes increasing the sampling probability for classes with a lower learning status, thus balancing the learning process more effectively. To demonstrate the efficacy of our proposed resampling strategy, we conducted comparisons with recent state-of-the-art methods, BiS ~\cite{he2021rethinking} and CReST~\cite{wei2021crest} with our resampling strategy only. Table.~\ref{tab:imbalance_compa_cls} and Table.~\ref{tab:imbalance_compa_det} indicate that our method still achieves better performance than BiS in both 3D detection and classification tasks when only the sampling part is utilized.

\subsection{Difference Between Our Method and FlexMatch}
FlexMatch~\cite{zhang2021flexmatch} is a semi-supervised learning method incorporating the idea of curriculum learning. Developed with the aim of enhancing the incorporation of unlabeled data during the initial training phases, FlexMatch~\cite{zhang2021flexmatch} employs a strategy known as 'threshold warm-up.' This tactic involves adjusting the threshold based on the extent of unused unlabeled data. Despite the adoption of this approach, a considerable portion of the unlabeled data remains underutilized during the training process due to the intricate nature of 3D data. This scenario leads to a reduction in the dynamic threshold for each class when integrating FlexMatch.  Moreover, FlexMatch adjusts each class's threshold based on the count of each class's pseudo-labels. While this strategy is effective for balanced datasets, it faces difficulties with commonly used 3D datasets~\cite{dai2017scannet,uy-scanobjectnn-iccv19,song2015sun,wu20153d}, where class-wise labeled data often follows a long-tail distribution This imbalance leads FlexMatch to set a much higher threshold for 'airplane' than for 'bowl'. As seen in Fig.\ref{fig:flex-threshold}, this approach results in considerably low and uneven thresholds, introducing substantial noise, particularly for minority classes, and ultimately leading to suboptimal performance in 3D tasks. In contrast, we dynamically adjusted thresholds according to class-level confidence. As shown in Fig.\ref{fig:our-threshold}, our methods have a more balanced dynamic threshold, even when faced with an imbalanced dataset. Consequently, our method optimizes the use of unlabeled data, minimizes noise introduction, and proves to be more adaptable than FlexMatch.

\begin{table}[t!]
\centering
\adjustbox{width=350pt,height=520pt,keepaspectratio=true}
{
	\begin{tabular}{|c|c|c|c|c|}

	    \hline
	     \multirowcell{2}{ } & 
	    \multicolumn{2}{c|}{SUN RGB-D 2\%} &
	    \multicolumn{2}{c|}{SUN RGB-D 5\%} \\ \cline{2-5}
	   & mAP $@$0.25 & mAP $@$0.5 & mAP $@$0.25 & mAP $@$0.5
        \\
        \hline
        Baseline &26.8 $\pm$ 1.1 &10.6 $\pm$ 0.5 & 39.7 $\pm$ 0.9 & 20.6 $\pm$ 0.7   \\
        \hline
        CReST~\cite{wei2021crest} + Baseline & 28.4 $\pm$ 0.9 & 11.3 $\pm$ 0.5 & 41.9 $\pm$ 0.8 & 22.7 $\pm$ 1.0  \\
        \hline
        BiS~\cite{he2021rethinking} + Baseline & 29.2 $\pm$ 0.7 & 11.5 $\pm$ 0.4 & 41.5 $\pm$ 1.1 & 22.4 $\pm$ 0.8   \\
        \hline
        Ours Resampling + Baseline & \textbf{31.5 $\pm$ 0.5}  &  \textbf{13.2 $\pm$ 0.6}  & \textbf{42.8 $\pm$ 0.4} &  \textbf{24.1 $\pm$ 0.7}    \\
        \hline
        
	\end{tabular}
}
 \caption{\small Comparative studies with state-of-the-art class imbalanced SSL method for 3D object detection.}
\label{tab:imbalance_compa_det}
\end{table}

\begin{table}[t!]
\centering
\adjustbox{width=320pt,height=480pt,keepaspectratio=true}
{
	\begin{tabular}{|c|c|c|c|c|}

	    \hline
	     \multirowcell{2}{ } & 
	    \multicolumn{2}{c|}{ModelNet40 5\%} &
	    \multicolumn{2}{c|}{ModelNet40 10\%} \\ \cline{2-5}
	   & Overall Acc & Mean Acc & Overall Acc & Mean Acc
        \\
        \hline
        Baseline & 78.9 & 71.1 & 85.5& 79.4     \\
        \hline
        CReST~\cite{wei2021crest} + Baseline & 80.2 & 71.9  & 86.0 & 80.7   \\
        \hline
        BiS~\cite{he2021rethinking} + Baseline & 79.7 & 72.3 & 86.1  &80.3 \\
        \hline
        Ours Resampling + Baseline & \textbf{81.9} & \textbf{73.8} & \textbf{87.7} & \textbf{82.2}     \\
        \hline
	\end{tabular}
}
 \caption{\small Comparative studies with state-of-the-art class imbalanced SSL method for 3D object classification.}
\label{tab:imbalance_compa_cls}
\end{table}

To provide a thorough understanding of our model's performance enhancement, we conduct a detailed analysis comparing the results of our model to those of FixMatch and FlexMatch. The average class-level confidence of our method, FixMatch, and FlexMatch of networks trained on ModelNet40 datasets are computed. In Fig.\ref{fig:modelnet_flexmatch_aveconfid}, an interesting observation emerges: despite FlexMatch's capability to boost confidence in some classes, the general class-level confidence remains imbalanced. This can be attributed to the fact that FlexMatch wasn't originally designed with the primary objective of addressing imbalanced datasets. Consequently, its capacity to recalibrate the learning status in these situations is limited. Conversely, as depicted in Figure \ref{fig:modelnet_our_aveconfid}, our proposed method exhibits significant improvements in class-level confidence. This notable enhancement is derived from our innovative approach, which integrates a dynamic threshold and a resampling strategy into the learning process. Through these integrated techniques, our method successfully elevates the average confidence levels across all classes, resulting in the achievement of a more balanced learning status. It's important to highlight that our method not only excels in enhancing confidence within specific classes but also addresses the broader challenge of class imbalance, thereby providing a comprehensive solution for improving overall learning performance.

\section{Conclusion}

This work presents DyConfidMatch, an innovative approach for 3D semi-supervised learning, addressing critical challenges such as data imbalance and varying class learning difficulties through dynamic thresholding and class-level re-sampling. Our extensive experiments demonstrate that DyConfidMatch surpasses existing state-of-the-art techniques in 3D object classification and detection tasks, highlighting its adaptability and efficacy across diverse datasets. One of the primary strengths of DyConfidMatch is its ability to dynamically balance learning across classes, leading to a more robust and equitable use of both labeled and unlabeled data. By leveraging class-level confidence to adjust thresholds and re-sample data, the method provides a nuanced understanding of the model’s performance, ensuring that even underrepresented classes are effectively learned. This adaptability makes DyConfidMatch a versatile tool that can be integrated into various semi-supervised learning frameworks.

However, the method is not without its limitations. The reliance on class-level confidence for dynamic thresholding, while effective in 3D contexts, does not translate as well to 2D tasks, indicating a potential area for further refinement. Additionally, the approach requires careful calibration of thresholds and re-sampling strategies to optimize performance, which may vary depending on the specific characteristics of the dataset. Future research should focus on extending DyConfidMatch to 2D datasets, where modifications to the learning status estimation and thresholding strategies could enhance its applicability. Incorporating advanced techniques for confidence estimation, such as uncertainty quantification, could further improve the method's robustness. Furthermore, applying DyConfidMatch to more complex, real-world scenarios like outdoor object detection or multi-modal data fusion would be essential for testing its generalizability and practical utility.

\section*{Acknowledgement}

This work was partially supported by the Simulation Based Reliability and Safety Program for modeling and simulation of military ground vehicle systems, under the technical services contract No. W56HZV-17-C-0095 with the U.S. Army DEVCOM Ground Vehicle Systems Center (GVSC) and was partially supported by the Automotive Research Center (ARC), a US Army Center of Excellence for modeling and simulation of ground vehicles, under Cooperative Agreement W56HZV-19-2-0001 with the US Army DEVCOM GVSC.



 \bibliographystyle{elsarticle-num} 
 \bibliography{refs}





\end{document}